\pdfoutput=1
\PassOptionsToPackage{table,dvipsnames}{xcolor}
\documentclass[11pt,letterpaper,logo,onecolumn]{main}
\usepackage[numbers,sort&compress,square]{natbib}

\usepackage{microtype}

\usepackage{enumitem}
\usepackage{needspace}

\usepackage{amsmath,amssymb,amsthm}
\usepackage{bm}

\usepackage{algorithm}
\usepackage[noend]{algpseudocode}

\usepackage{graphicx}
\usepackage{float}
\usepackage{wrapfig}
\usepackage{placeins}

\usepackage{booktabs}
\usepackage{multirow}
\usepackage{array}
\usepackage{tabularx}
\usepackage{fvextra}
\usepackage{siunitx}
\usepackage{etoolbox}
\usepackage[normalem]{ulem}
\usepackage{fontawesome}
\usepackage[most]{tcolorbox}
\definecolor{gblue9}{RGB}{23,78,166}

\newcolumntype{P}{S[table-format=2.1]}
\newcolumntype{O}{S[table-format=3.1]}
\newcolumntype{Q}{S[table-format=2.1,
                    table-space-text-post={\textsuperscript{\dag}}]}
\newcolumntype{D}{S[table-format=+2.1, print-implicit-plus]}

\robustify\bfseries
\robustify\uline
\newrobustcmd{\B}{\bfseries}                                    %
\newrobustcmd{\U}[2]{{\uline{\tablenum[table-format=#1]{#2}}}}  %

\newcolumntype{Y}[1]{>{\hsize=#1\hsize\raggedright\arraybackslash}X}

\AtBeginEnvironment{table}{\setlength{\belowcaptionskip}{5pt}}
\AtBeginEnvironment{figure}{%
  \setlength{\abovecaptionskip}{6pt}%
  \setlength{\belowcaptionskip}{0pt}}

\newcommand{\keepfinding}{\Needspace{0.13\textheight}}

\fancypagestyle{titlefolio}[firststyle]{%
  \fancyfoot[L]{\footerfont
                The full author list appears at the end of the paper.\quad
                Correspondence:
                \texttt{mingzhang23@m.fudan.edu.cn},
                \texttt{tgui@fudan.edu.cn}, \texttt{qz@fudan.edu.cn},
                \texttt{congzheng@tencent.com}, and
                \texttt{maxmpan@tencent.com}.}%
  \fancyfoot[R]{\footerfont\thepage}}

\fancypagestyle{authorfolio}[fancy]{%
  \fancyfoot[L]{\footerfont\textsuperscript{*}Equal contribution.\quad
                \textsuperscript{\dag}Corresponding authors.}}

\newcommand{\tnote}[1]{\par\vspace{2.5pt}\noindent\parbox{\linewidth}{%
  \footnotesize\setlength{\parindent}{0pt}#1}}

\newtcolorbox[auto counter]{diagnosticcase}[4][]{
  enhanced,
  breakable,
  pad at break*=2mm,
  colback=white,
  colframe=black!28,
  colbacktitle=black!6,
  colbacklower=black!2,
  coltitle=black,
  boxrule=0.55pt,
  arc=1.2mm,
  left=7pt,
  right=7pt,
  top=5pt,
  bottom=5pt,
  before skip=6pt,
  after skip=7pt,
  before={\par\Needspace{0.17\textheight}},
  fonttitle=\small,
  title={#4\hspace{0.35em}\textbf{Case~\thetcbcounter{} · #2}%
    \hfill\textnormal{\footnotesize #3}},
  #1
}
\newtcolorbox{casepanel}[1]{
  enhanced,
  breakable,
  colback=black!2,
  colframe=black!16,
  colbacktitle=black!7,
  coltitle=black,
  boxrule=0.4pt,
  arc=0.8mm,
  left=5pt,
  right=5pt,
  top=3pt,
  bottom=3pt,
  fonttitle=\footnotesize\bfseries,
  title={#1}
}
\newcommand{\casemeta}[4]{%
  \begin{tabularx}{\linewidth}{@{}Y{1.0}Y{0.9}Y{1.2}Y{0.9}@{}}
    \textcolor{black!55}{\scriptsize\textsf{ANSWERING MODEL(S)}} &
    \textcolor{black!55}{\scriptsize\textsf{TASK / RECORD}} &
    \textcolor{black!55}{\scriptsize\textsf{EVIDENCE}} &
    \textcolor{black!55}{\scriptsize\textsf{INTERACTION}} \\[-1pt]
    {\small #1} & {\small #2} & {\small #3} & {\small #4}
  \end{tabularx}\par\smallskip
}
\newtcolorbox[auto counter, number format=\Alph]{manualcard}[3][]{
  enhanced,
  breakable,
  colback=white,
  colframe=black!28,
  colbacktitle=black!6,
  coltitle=black,
  boxrule=0.55pt,
  arc=1.2mm,
  left=7pt,
  right=7pt,
  top=6pt,
  bottom=6pt,
  before skip=8pt,
  after skip=10pt,
  fonttitle=\small,
  title={\textbf{Manual~\thetcbcounter{} · #2}%
    \hfill\textnormal{\footnotesize #3}},
  title after break={\textbf{Manual~\thetcbcounter{} · #2}\
    \textnormal{(continued)}\hfill\textnormal{\footnotesize #3}},
  #1
}

\newcommand{\passstep}{\textcolor{ForestGreen}{\textsf{PASS}}}
\newcommand{\failstep}{\textcolor{BrickRed}{\textsf{FAIL}}}

\hypersetup{colorlinks=true, linkcolor=black, citecolor=blue, urlcolor=blue}
\usepackage{cleveref}
\crefname{tcb@cnt@diagnosticcase}{case}{cases}
\Crefname{tcb@cnt@diagnosticcase}{Case}{Cases}
\crefname{tcb@cnt@manualcard}{manual}{manuals}
\Crefname{tcb@cnt@manualcard}{Manual}{Manuals}

\usepackage{xspace}
\newcommand{\bench}{\textsc{ExplorationBench}\xspace}
\newcommand{\aliencode}{\textsc{AlienCode}\xspace}
\newcommand{\alienlogic}{\textsc{AlienLogic}\xspace}
\definecolor{AlienCodeClay}{HTML}{CC785C}
\definecolor{AlienCodeTint}{HTML}{F1E1D8}
\definecolor{AlienLogicWalnut}{HTML}{4F4234}
\definecolor{AlienLogicTint}{HTML}{E8E1D3}
\newcommand{\aliencodelogo}{%
  \raisebox{-0.27\height}{%
    \includegraphics[height=1.45em]{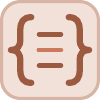}}}
\newcommand{\alienlogiclogo}{%
  \raisebox{-0.27\height}{%
    \includegraphics[height=1.45em]{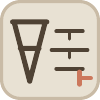}}}

\newcommand{\sandboxbandcols}[4]{%
  \multicolumn{#1}{@{}l@{}}{%
    \rlap{{\setlength{\fboxsep}{0pt}%
      \colorbox{#2}{\makebox[\linewidth]{%
        \rule[-0.32\baselineskip]{0pt}{1.28\baselineskip}#3\hfill#4}}}}}}

\newcommand{\brandicon}[1]{%
  \ifstrequal{#1}{kimi}{\raisebox{-0.12\height}{%
    \includegraphics[height=0.91em]{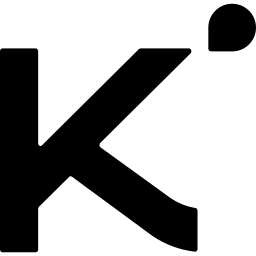}}}{}%
  \ifstrequal{#1}{bytedance-light}{\raisebox{-0.16\height}{%
    \includegraphics[height=1.00em]{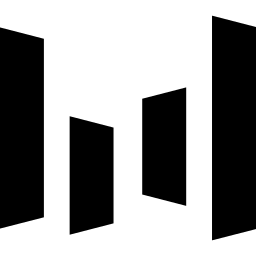}}}{}%
  \ifstrequal{#1}{claude-ai}{\raisebox{-0.18\height}{%
    \includegraphics[height=0.92em]{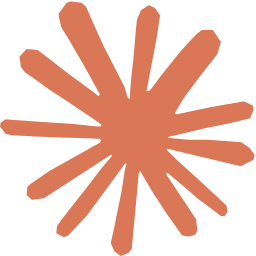}}}{}%
  \ifstrequal{#1}{deepseek-color}{\raisebox{-0.18\height}{%
    \includegraphics[height=0.92em]{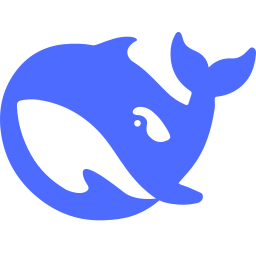}}}{}%
  \ifstrequal{#1}{gemini-color}{\raisebox{-0.18\height}{%
    \includegraphics[height=0.92em]{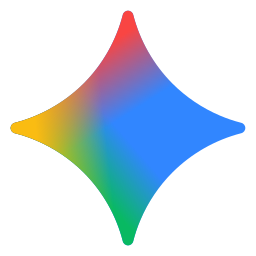}}}{}%
  \ifstrequal{#1}{grok}{\raisebox{-0.18\height}{%
    \includegraphics[height=0.92em]{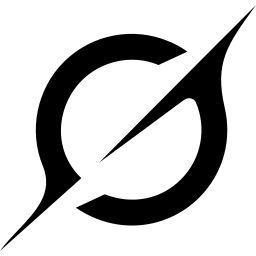}}}{}%
  \ifstrequal{#1}{hunyuan-color}{\raisebox{-0.18\height}{%
    \includegraphics[height=0.92em]{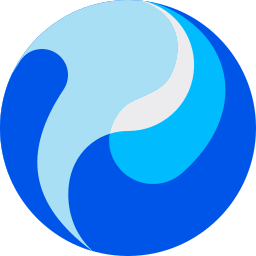}}}{}%
  \ifstrequal{#1}{openai}{\raisebox{-0.18\height}{%
    \includegraphics[height=0.92em]{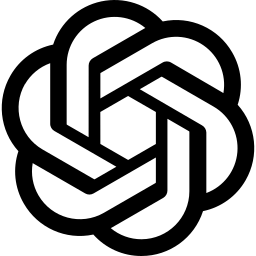}}}{}%
  \ifstrequal{#1}{qwen-color}{\raisebox{-0.18\height}{%
    \includegraphics[height=0.92em]{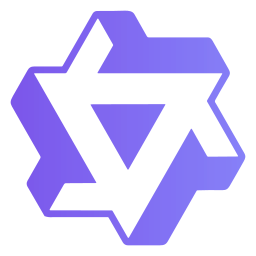}}}{}}
\newrobustcmd{\modelname}[2]{\mbox{\brandicon{#1}\hspace{0.32em}#2}}
\newrobustcmd{\aliencodebrand}{%
  \mbox{\raisebox{-0.18\height}{%
    \includegraphics[height=1.08em]{figures/mainfig/aliencode_mark.pdf}}%
    \hspace{0.28em}\textsc{AlienCode}}\xspace}
\newrobustcmd{\alienlogicbrand}{%
  \mbox{\raisebox{-0.18\height}{%
    \includegraphics[height=1.08em]{figures/mainfig/alienlogic_mark.pdf}}%
    \hspace{0.28em}\textsc{AlienLogic}}\xspace}
\newcommand{\findingcodeicon}{%
  \raisebox{-0.22\height}{%
    \includegraphics[height=1.35em]{figures/mainfig/aliencode_mark.pdf}}}
\newcommand{\findinglogicicon}{%
  \raisebox{-0.22\height}{%
    \includegraphics[height=1.35em]{figures/mainfig/alienlogic_mark.pdf}}}
\newcommand{\findingbothicons}{%
  \findingcodeicon\hspace{0.15em}\findinglogicicon}
\newcommand{\findingbox}[2]{%
  \begin{tcolorbox}[
    enhanced,
    colback=black!3,
    colframe=black!28,
    boxrule=0.55pt,
    arc=0.8mm,
    left=5pt,
    right=6pt,
    top=4pt,
    bottom=4pt,
    before skip=8pt,
    after={\par\nobreak\vspace{5pt}},
    drop shadow=black!16]
    \begin{tabularx}{\linewidth}{@{}>{\centering\arraybackslash}p{3.5em}X@{}}
      #1 & \textbf{#2}
    \end{tabularx}
  \end{tcolorbox}}

\newif\ifdraftresults
\draftresultsfalse
\ifdraftresults
  \newcommand{\result}[1]{\textcolor{violet}{#1}}
\else
  \newcommand{\result}[1]{#1}
\fi

\newcommand{\spec}[1]{#1}
\newcommand{\Hstate}{H_t}

\newcommand{\authorname}[1]{\textbf{#1}}

\DeclareRobustCommand{\titlelinks}{%
  \begingroup
    \hypersetup{urlcolor=black}%
    \normalsize\textsf{%
      \href{https://www.explorationbench.com}{\faGlobe~Website}%
      \qquad
      \href{https://github.com/Tencent-Hunyuan/ExplorationBench}{\faGithub~Code}%
      \qquad
      \href{https://www.explorationbench.com/blog/}{\faRss~Blog}}%
  \endgroup}

\newcommand{\myheaderbreak}{\\}
\title{\bench: Measuring AI Systems' \myheaderbreak
  Exploration in Verifiable Alien Worlds}

\author{
  \normalsize
  Fudan University
  \qquad
  Hunyuan Team, Tencent
  \qquad
  \textbf{Tsinghua University}\\
  \vspace{0.10cm}
  \titlelinks
}

\newcommand{\contributorlist}{%
  \authorname{Ming Zhang}\textsuperscript{*\dag},
  \authorname{Zhenghao Xiang}\textsuperscript{*},
  \authorname{Peizhong Gao}\textsuperscript{*},
  \authorname{Yujiong Shen},
  \authorname{Yuhui Wang},
  \authorname{Zhonghan Yue},
  \authorname{Shihan Dou},
  \authorname{Zhangyue Yin},
  \authorname{Junjie Ye},
  \authorname{Shichun Liu},
  \authorname{Weihuang Zheng},
  \authorname{Jiahao Chen},
  \authorname{Jiayi Chen},
  \authorname{Hongzhang Liu},
  \authorname{Jiaqi Shao}}
\newcommand{\leadlist}{%
  \authorname{Tao Gui}\textsuperscript{\dag},
  \authorname{Qi Zhang}\textsuperscript{\dag},
  \authorname{Xuanjing Huang},
  \authorname{Suncong Zheng}\textsuperscript{\dag},
  \authorname{Maxm Pan}\textsuperscript{\dag}}

\newcommand{\fullauthorlist}{%
  \section*{Full Author List}
  \contributorlist{}

  \vspace{0.35em}
  \par\noindent\leadlist{}%
  \thispagestyle{authorfolio}}

\begin{abstract}
Scientific discovery begins where known problems end. There, AI systems must
engage in exploration: framing hypotheses, designing experiments, and
iterating on the results. However, evaluating this ability is difficult: (1)
how to verify whether a genuinely new hypothesis holds, and (2) how to
determine whether a system has discovered it through exploration or merely
recalled related knowledge from pre-training data.
To this end, we introduce \bench{}, which turns the wicked problem of
evaluating scientific exploration into a concrete and tractable framework
built on verifiable Alien Worlds: their rules are executable, so every answer
can be checked exactly, and they conflict with familiar knowledge, so recall
alone cannot solve the tasks. The benchmark contains two sandboxes, \aliencode{}
(\spec{31} discovery targets, \spec{70} tasks) and \alienlogic{}
(\spec{24} discovery targets, \spec{70} tasks). Each sandbox provides a flawed
manual, task-specific environmental feedback, and a dedicated tool-call
schema. Systems use these resources to explore the sandbox, then solve held-out
tasks. We evaluate ten AI systems and find that the strongest systems can
acquire and apply unfamiliar rules, while performance varies substantially across trajectories and continued exploration can stall or reverse earlier gains.
\bench{} represents a step towards AI systems that can acquire and apply
genuinely new knowledge through exploration in unknown environments.
\end{abstract}

\begin{document}
\raggedbottom
\clubpenalty=10000
\widowpenalty=10000
\makeatletter\@clubpenalty=10000\makeatother
\setlength{\parskip}{0.3\baselineskip}
\maketitle
\thispagestyle{titlefolio}
\renewcommand{\myheaderbreak}{ }
\enlargethispage{0.4cm}

\begingroup
\setlength{\intextsep}{0pt}
\begin{figure}[H]
  \vspace{-2pt}
  \centering
  \includegraphics[width=\linewidth]{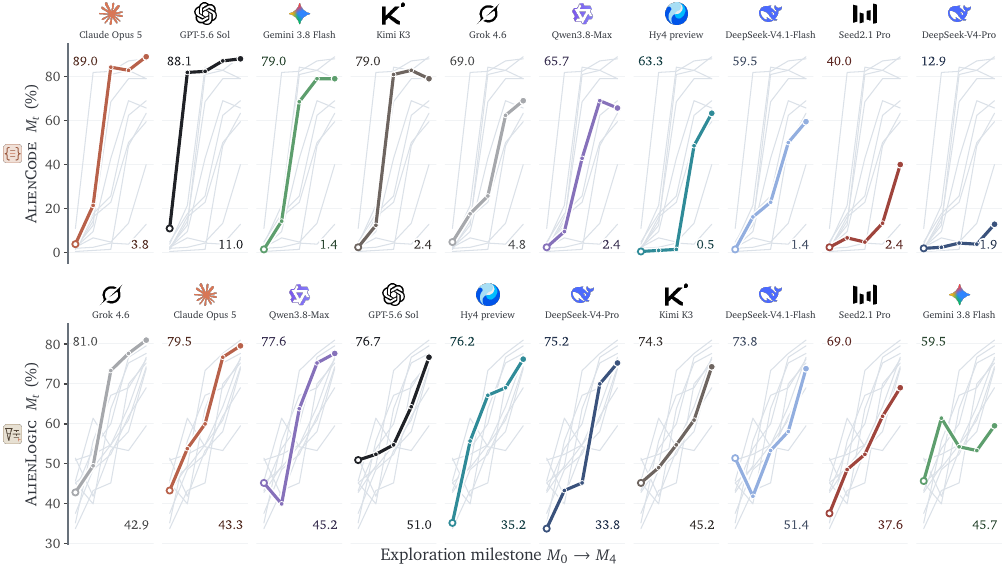}
  \caption{\textbf{Held-out accuracy over autonomous exploration.} Top:
  \aliencode{}; bottom: \alienlogic{}. Each cell highlights one system's
  $M_0$--$M_4$ curve; the other nine are grey. The upper-left and lower-right
  values are $M_4$ and $M_0$. Cells are ordered by $M_4$ within each sandbox.
  Each curve is the Best@$3$ trajectory, and each milestone is the mean of
  three answers per held-out question.}
  \label{fig:budget-curves}
\end{figure}
\endgroup
\newpage

\begingroup
\setlength{\intextsep}{0pt}
\begin{figure}[H]
  \centering
  \includegraphics[width=\linewidth]{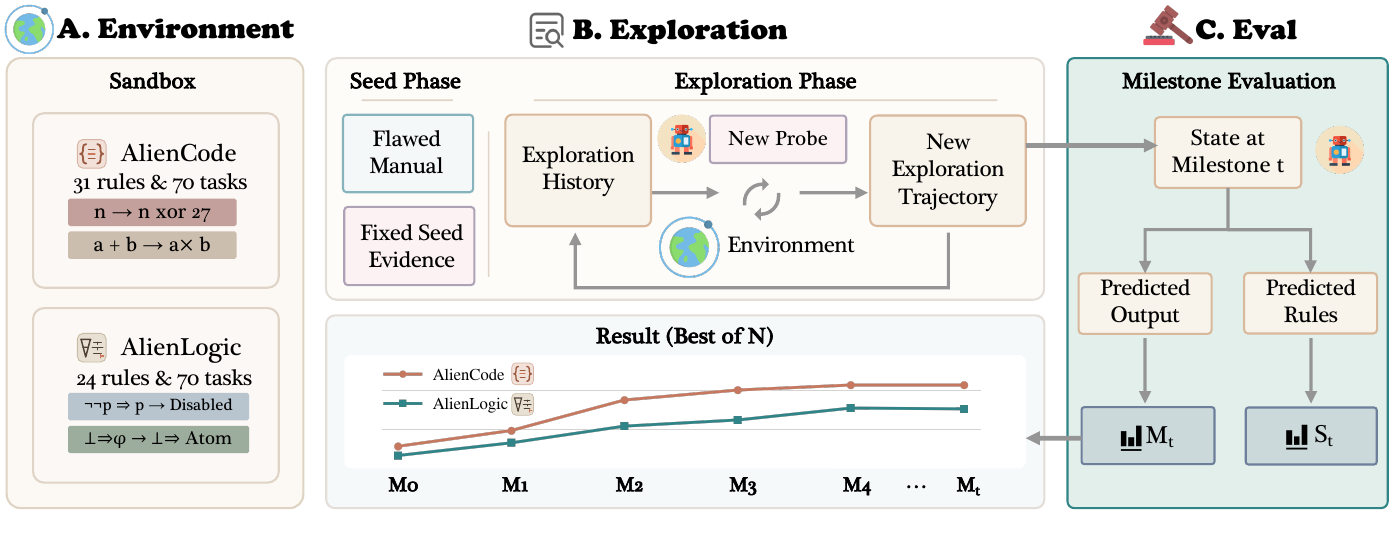}
  \caption{\textbf{\bench{}: what a system is given, what it may
  do about it, and how it is scored.}
  \textbf{(A)} Two executable sandboxes. \aliencode{} has \spec{31} discovery targets and \alienlogic{} \spec{24}, so the
  manual each ships with is wrong in ways no amount of prior knowledge
  recovers -- the changes have to be found from evidence.
  \textbf{(B)} Every system starts from the same flawed manual and the same
  fixed worked examples, then runs four rounds in which it chooses its own
  probes, runs them in the environment, and adds the results to its
  exploration history. Nothing else enters the context.
  \textbf{(C)} At each milestone $M_t$, the system is tested without tools: it
  answers the \spec{70} held-out tasks and separately reports the rules it
  believes it has found. A trajectory is scored on where it ends, and a system
  on its best complete trajectory.}
  \label{fig:overview}
\end{figure}
\endgroup

\section{Introduction}
\label{sec:intro}

Scientific research advances through exploration. Researchers begin with a
tentative understanding, choose what evidence to collect, revise their beliefs
in light of the outcomes, and apply what they learn to new problems. Current
large language models (LLMs) perform strongly on static evaluations of
knowledge, mathematics, and professional tasks
\citep{hendrycks2021mmlu,maa2026aime,patwardhan2025gdpval}, but these
evaluations mainly test whether a model can retrieve and reason with knowledge
it already has. Exploration asks for something different. A system must
decide what evidence to collect, accumulate that evidence across its
interaction history, distill new knowledge from the outcomes, and apply that
knowledge to new problems. As AI systems enter scientific and engineering
workflows \citep{wang2023scientific,chan2025mlebench}, measuring this ability
separately from recall and one-shot reasoning becomes increasingly important.

Evaluating exploration ability is difficult because two requirements are in
tension. The tasks must be new to the system, so that success cannot come
from knowledge acquired during pre-training, yet their answers must be fully
known to the evaluator, so that success can be verified
\citep{chollet2019measure}. Established domains such as mathematics, coding,
and factual question answering meet the second requirement but not the first.
A model may reproduce what it memorized, and contamination is hard to exclude
for black-box systems \citep{oren2024contamination}. Genuinely novel outputs,
such as a new mathematical result or scientific hypothesis, meet the first
requirement but not the second. Verifying them may require expert proof
checking, specialized experiments, or years of observation
\citep{wang2023scientific}, and without such verification an evaluator cannot
tell a genuine discovery from a plausible but incorrect claim.

Existing benchmarks cover parts of this setting (\cref{sec:related}).
Context-learning benchmarks place the new knowledge in the prompt
\citep{dou2026clbench,clbench2026continual,agarwal2024manyshot}, so the model
reads the evidence rather than collecting it. Interactive discovery benchmarks
let agents gather evidence in fictional worlds or through chosen experiments
\citep{tang2024mars,jansen2024discoveryworld,zheng2026newtonbench}, but they
score the state of that world or the inferred law itself. None of them asks
whether a system can collect the evidence it needs and then apply what it
learned to unseen tasks after the interaction has ended.

In this work, we introduce \bench{}, a benchmark that evaluates exploration in
verifiable alien worlds (\cref{fig:overview}). Each world is a deterministic,
executable environment whose hidden rules conflict with familiar semantics.
\aliencodebrand is a small programming language with \spec{31} discovery targets,
and \alienlogicbrand is a natural-deduction system with \spec{24} discovery targets. In \aliencode{}, for example, integer literals are silently
XOR-ed with 27, so \texttt{EMIT(100)} prints \texttt{127}, and \texttt{PLUCK}
counts positions from one although the manual says zero. A system starts from
this flawed manual and a few worked examples, then explores for four rounds by
submitting programs or proofs and reading the results. After each round it is
tested without tool access. It states the rules it believes hold and solves
\spec{70} held-out tasks, which an interpreter or a proof-checker grades
exactly.

\bench{} offers several properties that make exploration measurable.
(1)~\textbf{Resistant to recall.} The hidden rules contradict both the manual
and pre-training priors, so recalled knowledge misleads rather than helps.
Before exploring, no \aliencode{} trajectory exceeds \result{15.7\%}.
(2)~\textbf{Exactly verifiable.} Every answer is checked by executing it, so
grading needs no LLM judge, as in test-based agent evaluation
\citep{jimenez2024swebench,merrill2026terminalbench}.
(3)~\textbf{Resolved over the process.} Beyond the final score, the benchmark
records the probes a system chooses and, at every milestone, the rules it
reports and its held-out accuracy. This separates discovering a rule from
using it. (4)~\textbf{Controlled.} Matched conditions remove
environment feedback, replace the system's probes with a fixed sequence or with
its own best sequence, or supply the complete rule set
(\cref{sec:formulation,sec:metrics}).

We evaluate ten frontier systems, each with three independent exploration
trajectories per sandbox, and rank them by Best@3, the best final accuracy
among the three. Exploration produces the knowledge the tasks require. After
four rounds the best \aliencode{} trajectory reaches \result{89.0\%}, whereas
the same number of model turns without environment feedback leaves systems at
\result{0.5}--\result{11.0\%}. It also matters who designs the experiments. In
\aliencode{}, handing a system back its own best probes without letting it
choose them lowers accuracy for \result{9} of \spec{10} systems, and randomized
probes barely help. A system's exploration ability differs across tasks, and
its rank in one sandbox barely predicts its rank in the other (Spearman
\result{0.35}). Discovering a rule and using it also come apart. Two
off-by-one rules enter \spec{51} of the \spec{70} \aliencode{} tasks and
the largest accuracy jumps coincide with their discovery, yet tasks whose required rules a system states
correctly are still solved only \result{73.4\%} of the time. In
\alienlogic{}, being told the rules (\result{96}--\result{100\%}) beats every
system's own exploration. Finally, exploration is unreliable. Trajectories of
one system under one budget end up to \result{73.3} points apart, far more than
repeated answers to the same questions vary, and \result{6} of the \result{30}
\aliencode{} trajectories end at least \result{3} points below an earlier
milestone.

More findings and case studies are presented in \cref{sec:experiments} and
\cref{app:cases}. Frontier systems can acquire unfamiliar rules through
exploration, but they do so unreliably and do not always use what they find.
\bench{} provides a testbed for measuring how AI systems acquire and apply new
knowledge, and for developing more reliable exploration methods.

\section{Related Work}
\label{sec:related}

In this section, we position \bench relative to three lines of work:
interactive scientific discovery and rule induction, context learning and
test-time adaptation, and verifiable agent evaluation
\citep{tang2024mars,jansen2024discoveryworld,dou2026clbench,dou2025evalearn,
jimenez2024swebench,yao2024taubench}.

\paragraph{Scientific discovery and interactive rule induction.}
Interactive benchmarks ask agents to infer hidden or altered rules from
evidence, connecting to causal world models and active system identification
\citep{lake2017building}. MARS and DiscoveryWorld study
investigation in fictional worlds
\citep{tang2024mars,jansen2024discoveryworld}, while NewtonBench and
SciLaws-Bench target scientific law discovery
\citep{zheng2026newtonbench,huang2026scilaws}. SciLaws-Bench pairs fixed real
observations with queryable parallel worlds, whose newly synthesized hidden
laws separate discovery from memorization.
Nearby settings measure compliance with stated constraints
in COLLIE \citep{yao2024collie} and search over experiment configurations in
MLAgentBench \citep{huang2024mlagentbench}. Voyager instead evaluates
open-ended skill accumulation in a sandbox \citep{wang2023voyager}.
\bench measures the complete path from selecting probes, through reporting
discoveries, to using them on unseen tasks. Its unfamiliar executable worlds
help separate knowledge acquired during evaluation from pre-trained recall.

\paragraph{Context learning and test-time adaptation.}
A growing position emphasizes experience generated by the system itself
\citep{silver2025experience}. CL-bench and CL-bench Life study learning from
complex provided contexts \citep{dou2026clbench,dou2026clbenchlife}, while
EvaLearn studies experience across sequential problems
\citep{dou2025evalearn}. SE-Bench moves adaptation into model weights
\citep{yuan2026sebench}, and EdgeBench examines longer-horizon learning in
real-world environments \citep{zhu2026edgebench}. Surveys organize
self-evolving agents across changes to weights, memory, tools, and prompts
\citep{gao2025selfevolving,fang2025comprehensive}. This setting also connects
to context engineering and in-context learning
\citep{mei2025contextengineering,brown2020gpt3}, with mechanisms studied
through induction circuits and implicit Bayesian inference
\citep{olsson2022induction,xie2022bayesian} and in-context state represented
explicitly in language-agent architectures \citep{sumers2023cognitive}.

Existing approaches typically study learning from supplied demonstrations
\citep{min2022rethinking}, feedback on a system's own attempts
\citep{shinn2023reflexion,madaan2023selfrefine,chen2024selfdebug}, or stored
trajectories replayed as context \citep{zhao2024expel}. Longer reasoning instead
spends inference without adding evidence \citep{wei2022cot}. \bench holds
parameters fixed and places these routes in one protocol. Autonomous exploration
selects probes online, hindsight exploration replays the system's own best
trajectory, and without-tool answering adds deliberation without environment
feedback. This separates who
selects evidence from whether evidence arrives.

\paragraph{Verifiable interactive evaluation.}
Verifiable agent evaluation grades outcomes rather than descriptions, building
on the interleaving of reasoning and tool use in ReAct \citep{yao2023react}.
SWE-bench grades repository patches with tests \citep{jimenez2024swebench},
WebArena scores website end states \citep{zhou2024webarena}, and $\tau$-bench
compares database states with annotated goals
\citep{yao2024taubench}. $\tau^2$-bench extends this setting to environments in
which both parties act \citep{barres2025tau2}. \bench shares this preference for
executable outcomes. An interpreter grades \aliencode{}, while a proof-checker
and bounded certifier grade \alienlogic{}. A related line turns the environment
itself into the prediction target: Qwen-AgentWorld trains a language world model
to simulate how an environment would respond, scored against recorded
observations across seven domains \citep{zuo2026qwen}. That asks how faithfully
a system can reproduce known dynamics, whereas \bench asks whether it can
uncover dynamics nobody stated. Here, the governing rules
must be discovered before they are used on closed-book tasks.
\Cref{tab:benchmark-comparison} summarizes the differences in evidence,
novelty, progress measurement, and transfer.

\section{ExplorationBench}
\label{sec:benchmark}

This section defines the exploration task and its experimental conditions,
describes the two alien worlds, and then defines the metrics. Exploration is
treated as three connected parts, probe selection, reported discovery, and
rule use, with no parameter updates.

\subsection{The task}
\label{sec:formulation}
\label{sec:design}

\paragraph{Setup.}
An episode is a tuple
$(\mathcal E,\mathcal R,\mathcal M,\mathcal D,\mathcal T)$. $\mathcal E$ is a
hidden, deterministic sandbox governed by a perturbed rule set $\mathcal R$.
$\mathcal M$ is a public, flawed manual. It describes the \emph{standard}
semantics, which are false for the perturbed parts of $\mathcal E$.
$\mathcal D$ is a fixed set of worked examples shared by every system.
$\mathcal T$ is an unseen task set held out from exploration. The system sees $\mathcal M$ and $\mathcal D$ and may interact with $\mathcal E$, but never
sees $\mathcal R$. A system that trusts the manual or its pre-training priors is
therefore wrong on exactly the parts of $\mathcal E$ that matter.

\paragraph{The exploration policy.}
Let $h_t=(\mathcal M,\mathcal D,x_{<t},f_{<t})$ denote the manual, the
worked examples, and the interaction history so far. At step $t$, the
system submits a tool input according to
\begin{equation}
  x_t\sim\pi(\,\cdot\mid h_t\,),\qquad f_t=\mathcal E(x_t).
  \label{eq:exploration-policy}
\end{equation}
We call each executable tool input $x_t$ a \emph{probe}. In \aliencode{} a
probe is a candidate program and its feedback $f_t$ is the exact program
output. In \alienlogic{} it is a candidate proof and $f_t$ is a compact
verifier result. The system uses this feedback to revise a hypothesis state
$H_t$ about $\mathcal R$ within a common maximum budget $B_{\max}$. $H_t$ and
the evidence in $h_t$ remain inside one context window, with no weight
updates, external or persistent memory, or scalar reward.

\paragraph{Protocol.}
\Cref{alg:protocol} gives the protocol. Before $M_0$, every system receives
the same worked examples, each a task with a reference program or proof and
the output the environment computes for it, and makes no tool calls. $M_0$
therefore follows identical evidence, not merely an identical opportunity to
collect it. The system then explores for four rounds. In each round it issues
tool calls whose arguments are programs or proofs, and the environment executes
them and returns deterministic feedback. At each milestone the system is tested
in a separate copy of the conversation with tools disabled. It reports the
rules it believes hold, $S_t$, and answers the held-out tasks. The copy is then
discarded, so testing never adds evidence to the exploration.

\paragraph{Exploration conditions.}
Five conditions vary whether environment feedback arrives and who directs it.
\emph{Autonomous exploration} ($a$) selects probes from the current history.
\emph{Hindsight exploration} ($h$) replays the probes of the same system's
Best@3 trajectory, chosen after the fact. \emph{Fixed-probe exploration} ($f$)
issues one model-independent probe sequence, the same for every system.
\emph{Without-tool answering} ($w$) adds model turns without environment
feedback, and \emph{direct answering} ($d$) adds none. A sixth condition,
\emph{open-book answering} ($o$), supplies the complete rule set, either before
exploration (O@$M_0$) or after autonomous exploration (A4+O). The three
feedback conditions use the same tool-calling interface, so they differ only in
who directs the probes. Conditions are compared by their $M_4$ endpoints,
Best@$n$ for autonomous exploration and the single trajectory each control
runs per system. Set against $M_4$, O@$M_0$ and A4+O separate finding the rules
from using them. Full definitions appear in \cref{app:controls}.

\subsection{The Alien Worlds}
\label{sec:sandboxes}

\begin{wrapfigure}[16]{r}{0.42\linewidth}
  \vspace{-1.5\baselineskip}
  \centering
  \includegraphics[width=\linewidth]{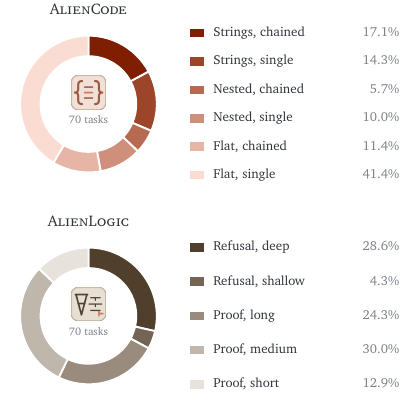}
  \caption{\textbf{Held-out task composition.} The \spec{70} tasks of each
  sandbox by family.}
  \label{fig:task-composition}
  \vspace{-0.5\baselineskip}
\end{wrapfigure}
\aliencode{} and \alienlogic{} share the task and closed-book evaluation
above. One is built on program semantics, the other on formal inference. Both
are deterministic and executable, and both deliberately conflict with familiar
priors. Some rules remain unchanged as red herrings, and every altered rule has
at least one held-out task. An interpreter checks \aliencode{} programs and a
proof-checker checks \alienlogic{} proofs, so scoring is deterministic, exact,
and free of LLM judges. Construction and verifier details are in
\cref{app:protocol,app:formal-metrics}.

\paragraph{\texorpdfstring{\aliencodebrand}{AlienCode}.}
This sandbox is a small calculation language whose familiar-looking operators
follow hidden semantics. It contains \spec{31} discovery targets and
\spec{70} held-out tasks, namely \spec{37} base tasks, \spec{15} nested
composition tasks, \spec{8} ceiling tasks, and \spec{10} rule-coverage tasks.
Together they exercise all \spec{31} discovery targets. Before evaluation, each
task is classified by representation (flat values, nested containers, or
multi-character strings) and by compositional depth (single-step,
single-algorithm, or multi-stage, \cref{fig:task-composition}). Programs are graded by an interpreter on private
evaluator inputs, so success requires rule-aware executable behavior rather
than memorizing displayed examples. During exploration, the environment tool
executes exactly the candidate program submitted and returns its output.

\paragraph{\texorpdfstring{\alienlogicbrand}{AlienLogic}.}
This sandbox is a first-order natural-deduction system with \spec{24} discovery targets, each a patched inference rule, and \spec{70} held-out tasks. A proof-checker verifies every
submitted proof, and designated unprovable tasks receive credit only when the system correctly declines to prove them. Certifier bounds and rule-side
conditions are given in \cref{app:protocol}. During exploration, the
environment tool checks candidate proofs and returns a compact verifier result.

\subsection{Metrics}
\label{sec:metrics}

\paragraph{What we measure.}
Held-out task performance is
$Q(H_t)=\Pr_{\tau\sim\mathcal{T}}\!\left[\text{the system solves }\tau\text{
given }H_t\right]$. The exploration curve $(Q(H_t))_{t=0}^{N}$ is reported as
$(M_0,\dots,M_N)$, with $M_0$ measured after the worked examples and $M_t$
after round $t$. The benchmark records three connected parts of the process
separately. \emph{Probe selection} is which probes $\pi$ selects, read from the
recorded probes and budget and compared across the exploration conditions. \emph{Reported discovery} is the rule set $S_t$ the system reports from $H_t$. \emph{Rule use} is whether the system can apply
$H_t$ to unseen tasks, measured by held-out accuracy. The three diverge in our
experiments (\cref{sec:experiments}), so they are reported separately rather
than as one score.

\paragraph{Held-out accuracy.}
Every held-out question is answered three times, each time in a fresh
tool-disabled copy of the conversation at that milestone. For trajectory $r$ at
milestone $t$, with $y^{(k)}_{r,\tau,t}\in\{0,1\}$ the verdict on the $k$-th
answer to task $\tau$,
\begin{equation}
  M_{r,t}=\frac{100}{|\mathcal T|}\sum_{\tau\in\mathcal T}
    \frac{1}{3}\sum_{k=1}^{3} y^{(k)}_{r,\tau,t}.
  \label{eq:milestone-accuracy}
\end{equation}
An answer that is missing or exhausts its time budget scores zero. Writing
$M^{(k)}_{r,t}$ for the score of the $k$-th pass alone, the \emph{answering
noise} $\sigma_{r,t}$ is the standard deviation of $M^{(1)}_{r,t}$,
$M^{(2)}_{r,t}$, and $M^{(3)}_{r,t}$. It measures how much the score moves when
the same knowledge answers again, with exploration held fixed.

\paragraph{Model score.}
Each system runs $n$ independent trajectories under the same protocol. The
primary score is the best endpoint reached by one complete trajectory,
\begin{equation}
  \operatorname{Best@}n
  = \max_{r\in\{1,\ldots,n\}} M_{r,4},
  \qquad
  r^\star=\min\arg\max_r M_{r,4},
  \label{eq:best-of-n}
\end{equation}
where the lowest trajectory index breaks an exact tie. The milestone curve,
rule reports, and budget reported with the headline score all come from
$r^\star$, and no synthetic trajectory is assembled from different trajectories
at different milestones or tasks. Beside Best@$n$ we report
$\operatorname{Mean@}n=n^{-1}\sum_r M_{r,4}$, the lowest endpoint, and every
trajectory, and rankings compare systems only at the same $n$.

\paragraph{Rule reports.}
At every milestone the system also states the rule set $S_t$ it currently
believes. In \aliencode{}, each of the \spec{31} evaluator-side rules is judged
stated correctly or not, and we report the number stated correctly
(\cref{eq:rule-report}). Each task $\tau$ exercises a known set of rules
$\mathcal R(\tau)$. The task is \emph{covered} at milestone $t$ when every rule
in $\mathcal R(\tau)$ is stated correctly, which splits held-out accuracy by
what the report says the system knows. \alienlogic{} has no rule-report score,
and its auxiliary diagnostic is correct refusal on unprovable theorems. Rule
reports are diagnostic and never enter $M_t$.

\paragraph{Dynamics and budget.}
We describe how a trajectory reaches its endpoint by its per-round steps and
its retained gain $G_r=M_{r,4}-M_{r,0}$ (\cref{eq:retained-gain}). The
benchmark also records the exploration budget $\mathbf B=(C,P,T)$ of tool
calls, probe units, and exploration tokens. The budget is reported (\cref{tab:budget})
but not scored, and its accounting is given in \cref{app:controls}.

\section{Results and Findings}
\label{sec:experiments}

\newcommand{\MainCodeRows}{%
\modelname{claude-ai}{Claude Opus 5} & max & 3.8 & 89.0 & 73.5 & 49.0 & 85.2 & 92.9 & 27 \\
\modelname{openai}{GPT-5.6 Sol} & max & 11.0 & 88.1 & 79.5 & 73.8 & 69.5 & 91.9 & 28 \\
\modelname{gemini-color}{Gemini 3.8 Flash} & high & 1.4 & 79.0 & 39.7 & 6.2 & 79.5 & 88.1 & 28 \\
\modelname{kimi}{Kimi K3} & max & 2.4 & 79.0 & 38.7 & 5.7 & 53.8 & 82.9 & 28 \\
\modelname{grok}{Grok 4.6} & xhigh & 4.8 & 69.0 & 53.2 & 37.1 & 52.4 & 79.0 & 27 \\
\modelname{qwen-color}{Qwen3.8-Max} & max & 2.4 & 65.7 & 64.0 & 61.4 & 42.4 & 69.5 & 23 \\
\modelname{hunyuan-color}{Hy4 preview} & high & 0.5 & 63.3 & 30.2 & 1.4 & 28.6 & 66.2 & 26 \\
\modelname{deepseek-color}{DeepSeek-V4.1-Flash} & max & 1.4 & 59.5 & 46.3 & 23.3 & 40.5 & 68.6 & 25 \\
\modelname{bytedance-light}{Seed2.1 Pro} & high & 2.4 & 40.0 & 24.8 & 4.8 & 62.9 & 56.7 & 26 \\
\modelname{deepseek-color}{DeepSeek-V4-Pro} & max & 1.9 & 12.9 & 7.3 & 2.4 & 39.0 & 52.9 & 22 \\
}
\newcommand{\MainLogicRows}{%
\modelname{grok}{Grok 4.6} & xhigh & 42.9 & 81.0 & 72.1 & 65.7 & 100.0 & 100.0 & 89.9 \\
\modelname{claude-ai}{Claude Opus 5} & max & 43.3 & 79.5 & 75.9 & 69.5 & 96.7 & 97.6 & 43.5 \\
\modelname{qwen-color}{Qwen3.8-Max} & max & 45.2 & 77.6 & 74.3 & 71.9 & 99.5 & 95.7 & 58.0 \\
\modelname{openai}{GPT-5.6 Sol} & max & 51.0 & 76.7 & 75.1 & 72.4 & 95.7 & 97.6 & 47.8 \\
\modelname{hunyuan-color}{Hy4 preview} & high & 35.2 & 76.2 & 69.2 & 61.4 & 98.6 & 96.7 & 59.4 \\
\modelname{deepseek-color}{DeepSeek-V4-Pro} & max & 33.8 & 75.2 & 66.3 & 61.4 & 97.6 & 95.2 & 59.4 \\
\modelname{kimi}{Kimi K3} & max & 45.2 & 74.3 & 69.4 & 61.4 & 98.1 & 99.0 & 65.2 \\
\modelname{deepseek-color}{DeepSeek-V4.1-Flash} & max & 51.4 & 73.8 & 60.6 & 52.9 & 98.1 & 95.2 & 44.9 \\
\modelname{bytedance-light}{Seed2.1 Pro} & high & 37.6 & 69.0 & 59.8 & 53.3 & 96.2 & 95.7 & 52.2 \\
\modelname{gemini-color}{Gemini 3.8 Flash} & high & 45.7 & 59.5 & 59.0 & 58.6 & 100.0 & 100.0 & 52.2 \\
}

Results are organized around the exploration milestones $M_0$ through $M_4$.
Each sandbox is scored on its \spec{70} held-out tasks, and every task is
answered three times at each milestone, so a score is a mean of three answers
rather than one.

\subsection{Setup}
\label{sec:setup}

\paragraph{Systems and trajectories.}
We evaluate ten frontier systems in each sandbox, each at the highest
reasoning setting its API offers. Every system runs $n{=}3$ independent
trajectories per sandbox. A trajectory is one continuous exploration history:
it starts from the shared worked examples, runs four exploration rounds, and is
scored at milestones $M_0$ through $M_4$ with the metrics of
\cref{sec:metrics}.

\subsection{Findings}
\label{sec:findings}

Eight findings address three questions. RQ1 asks whether systems can acquire
an alien world by exploring it, RQ2 how discovering a rule relates to using
it, and RQ3 how reliably exploration succeeds. Each system is scored by Best@3
(\cref{eq:best-of-n}), reported beside Mean@3 and all three trajectories. Two
open-book conditions supply the complete rule set, one before exploration
(O@$M_0$) and one after autonomous exploration (A4+O). Each control condition
is run once per system. \Cref{tab:main} lists each system's Best@3 trajectory
beside these references.

\begin{table}[!t]\centering
  \caption{\textbf{Held-out accuracy under autonomous exploration.} $M_0$ and
  $M_4$ of each system's Best@3 trajectory (\cref{eq:best-of-n}), the mean and
  the lowest $M_4$ over its three trajectories, and accuracy under open-book answering, with the complete rule set supplied
  before exploration (O@$M_0$) or after the Best@3 trajectory's
  exploration (A4+O). All values are percentages, each the
  mean of three answers per question; rows are ordered by Best@3 $M_4$.}
  \label{tab:main}
  \small
  \renewcommand{\arraystretch}{0.96}
  \setlength{\tabcolsep}{3.0pt}
  \begin{tabular*}{\linewidth}{@{\extracolsep{\fill}}ll PP PP OO P@{}}
    \toprule
    & & \multicolumn{2}{c}{Best@3 trajectory}
      & \multicolumn{2}{c}{Three trajectories}
      & \multicolumn{2}{c}{Open-book}
      & \multicolumn{1}{c@{}}{Diagnostic} \\
    \cmidrule(lr){3-4}\cmidrule(lr){5-6}\cmidrule(lr){7-8}\cmidrule(l){9-9}
    System & Reasoning & {$M_0$} & {\textbf{$M_4\uparrow$}} & {Mean@3}
      & {Worst} & {O@$M_0$} & {A4+O} & {Aux.} \\
    \midrule
    \sandboxbandcols{9}{AlienCodeTint!42!white}{%
      \aliencodelogo\hspace{0.45em}\textbf{\aliencode}
      \quad\textcolor{black!55}{\textit{program synthesis}}}{%
      \footnotesize\textcolor{black!60}{Aux. Rules $(/\spec{31})$}} \\
    \addlinespace[1pt]
    \MainCodeRows
    \addlinespace[2.5pt]
    \sandboxbandcols{9}{AlienLogicTint!42!white}{%
      \alienlogiclogo\hspace{0.45em}\textbf{\alienlogic}
      \quad\textcolor{black!55}{\textit{formal proof}}}{%
      \footnotesize\textcolor{black!60}{Aux. UnprovRec (\%)}} \\
    \addlinespace[1pt]
    \MainLogicRows
    \bottomrule
  \end{tabular*}
  \tnote{Reasoning labels denote API-specific request configurations; each
    system uses its highest available setting, held fixed across trajectories, sandboxes, and conditions. The labels are not a common compute scale. The
    diagnostic column differs by sandbox and the two are not comparable:
    \textsf{Rules} counts the hidden rules the Best@3 trajectory reports
    correctly at $M_4$, and \textsf{UnprovRec} is its rate of correct refusal on
    unprovable theorems. Neither enters $M_4$.}
\end{table}

\Needspace{0.16\textheight}
\subsubsection{RQ1. Can AI systems acquire an alien world through
exploration?}

\begin{figure}[!b]
  \centering
  \includegraphics[width=\linewidth]{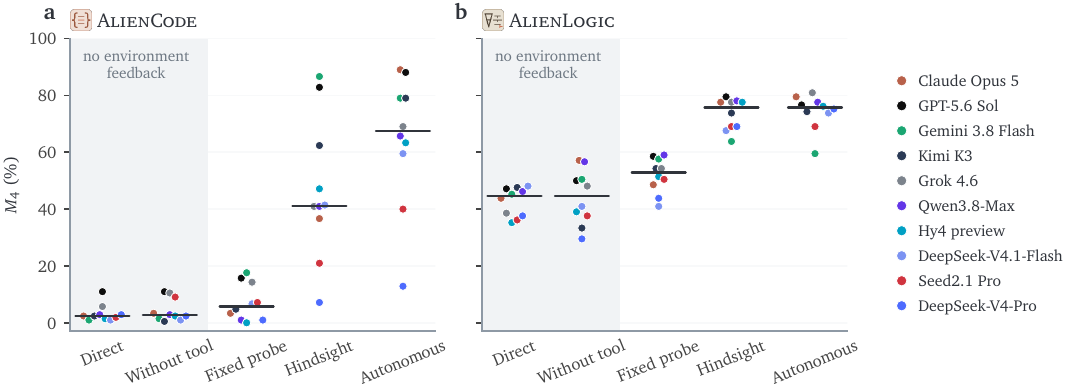}
  \caption{\textbf{Held-out accuracy at $M_4$ under five exploration
  conditions.} Each dot is one system; the horizontal line in each column
  marks the median over the ten systems. Autonomous exploration is shown at
  each system's Best@3 trajectory; hindsight exploration replays the probes of
  that same trajectory, and fixed-probe exploration issues model-independent
  probes at matched volume. Shaded columns receive no environment feedback, and from left to right the
  system takes a larger part in designing the experiments. Control
  conditions carry one trajectory per system, and every value is the mean of
  three answers per question.}
  \label{fig:conditions}
\end{figure}

\findingbox{\findingbothicons}{Finding 1. Exploration, not recall or thinking alone,
produces the knowledge the tasks require.}
Before exploration, recalled knowledge solves almost nothing in \aliencode{},
and no trajectory exceeds \result{15.7\%} at $M_0$. After four rounds, Best@3
reaches \result{89.0\%}, and \result{7} of \spec{10} systems exceed
\result{60\%}. \alienlogic{} starts higher, at \result{33.8}--\result{51.9\%},
because its rule changes leave part of standard natural deduction intact, and
Best@3 rises to \result{59.5}--\result{81.0\%}. Additional model turns do not
substitute for probing (\cref{fig:conditions,case:nofeedback}). Without-tool
answering adds the same model turns without environment feedback. It leaves
\aliencode{} at \result{0.5}--\result{11.0\%}, and in \alienlogic{} it changes
accuracy by \result{$-14.3$} to \result{$+13.3$} points relative to direct
answering, lowering it for \result{three} systems.

\keepfinding
\findingbox{\findingbothicons}{Finding 2. Exploration works best when the system
designs its own experiments.}
The three feedback conditions use the same tool-calling interface and differ
only in who designs the probes (\cref{fig:conditions}). Under autonomous
exploration the system designs each experiment from its own history. Hindsight
exploration hands back exactly the experiments of its Best@3 trajectory. The
system still reads every result and records its hypotheses, but it no longer
decides what to test next. Fixed-probe exploration samples experiments from
grammar-valid templates with a fixed seed, the same sequence for every system.
In \aliencode{}, the median falls from \result{67.4\%} under autonomous
exploration to \result{41.2\%} under hindsight and \result{5.7\%} under fixed
probes. Autonomous exploration beats hindsight for \result{9} of \spec{10}
systems, by a median of \result{17.4} points, although hindsight replays the
probes of the best of the three trajectories. The sandbox is deterministic, so
the evidence is identical, and the gap reflects designing the experiments
rather than receiving their results. Gemini 3.8 Flash is the exception
(\result{86.7\%} under hindsight against \result{79.0\%}). In \alienlogic{},
designing the experiments adds nothing. Autonomous and hindsight exploration
tie (median difference \result{0.2} points), and hindsight exceeds fixed probes
by a median of \result{22.1} points, so what matters there is which proofs are
tried rather than who chose them. Each hindsight run is a single trajectory and
varies on its own. Three replays of Hy4 preview's \aliencode{} probes end at
\result{47.1\%}, \result{10.0\%}, and \result{46.7\%}.

\begin{figure}[!tb]
  \centering
  \includegraphics[width=0.72\linewidth]{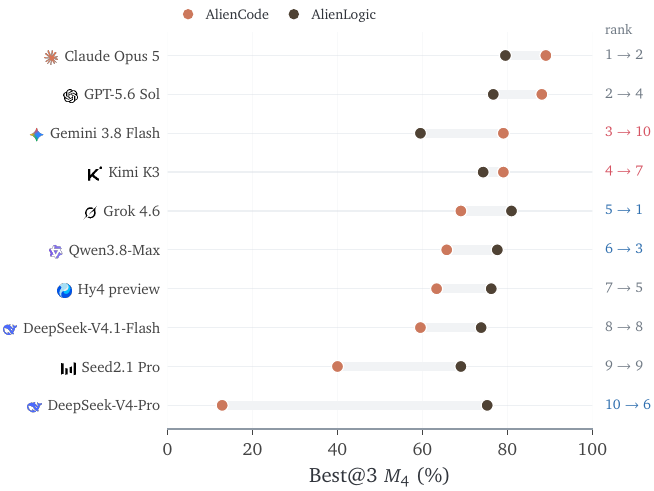}
  \caption{\textbf{Best@3 $M_4$ in the two sandboxes.} Each row is one system,
  ordered by its \aliencode{} score; the right column gives its rank in
  \aliencode{} and then in \alienlogic{}. Red marks a fall of three or more
  places, blue a rise of three or more.}
  \label{fig:sandbox-ranks}
\end{figure}

\keepfinding
\findingbox{\findingbothicons}{Finding 3. A system's exploration ability
varies across task settings.}
The Spearman correlation between the two Best@3 rankings is \result{0.35}
(\cref{fig:sandbox-ranks}).
Grok 4.6 ranks \result{fifth} in \aliencode{} and \result{first} in
\alienlogic{}, Gemini 3.8 Flash ties for \result{third} and ranks
\result{last}, and DeepSeek-V4-Pro moves from \result{last} to \result{sixth}.
The sandboxes also separate systems differently. \aliencode{} spreads Best@3 over
\result{12.9}--\result{89.0\%}, whereas \result{eight} of \spec{10} systems
fall within \result{73.8}--\result{81.0\%} in \alienlogic{}.

\Needspace{0.16\textheight}
\subsubsection{RQ2. How does discovering a rule relate to using it?}

\begin{figure}[!tb]
  \centering
  \includegraphics[width=\linewidth]{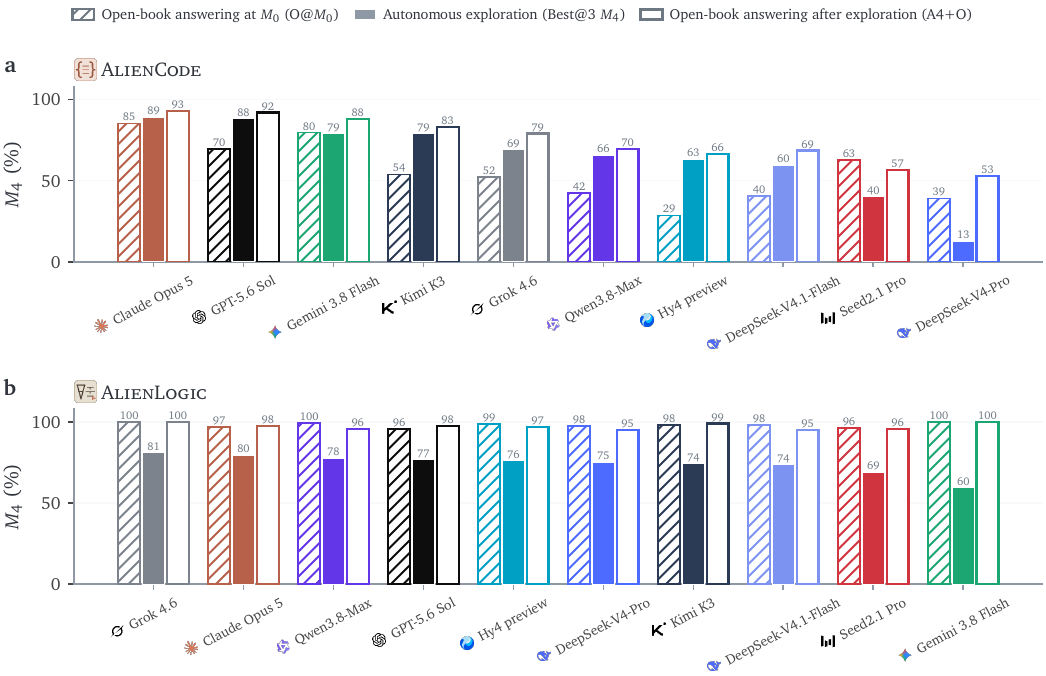}
  \caption{\textbf{Finding out versus being told.} Held-out accuracy at $M_4$
  under open-book answering before exploration (O@$M_0$, hatched),
  after autonomous exploration alone (Best@3, solid), and under open-book
  answering after that exploration (A4+O, outlined).
  Systems are ordered by Best@3 $M_4$ within each sandbox.}
  \label{fig:disclosure}
\end{figure}

\begin{figure}[!tb]
  \centering
  \includegraphics[width=\linewidth]{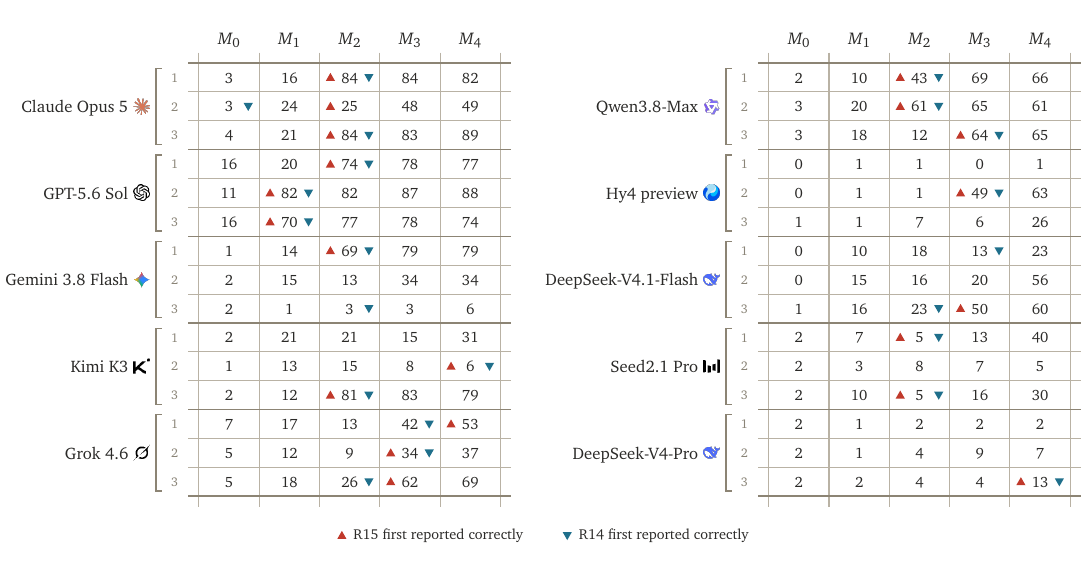}
  \caption{\textbf{Accuracy jumps when the keystone rules are found.} Each
  block is one system and each row one of its three \aliencode{} trajectories;
  columns are milestones, and each cell gives held-out accuracy (\%).
  Triangles mark the milestone at which the trajectory's rule report first
  states R15 (\texttt{CARVE} slice shift, pointing up) and R14
  (\texttt{PLUCK} index shift, pointing down) correctly.}
  \label{fig:keystone}
\end{figure}

\findingbox{\findingbothicons}{Finding 4. Discovering the rules beats being
told them in \aliencode{}, but not in \alienlogic{}.}
In \aliencode{} (\cref{fig:disclosure}), the Best@3 trajectory outscores
O@$M_0$ for \result{7} of \spec{10} systems, and Mean@3 does so for
\result{5}. GPT-5.6 Sol reaches \result{88.1\%} against \result{69.5\%} under
O@$M_0$, and Hy4 preview \result{63.3\%} against \result{28.6\%}. A4+O beats
O@$M_0$ in \result{26} of \result{30} trajectories, by a median of
\result{14.5} points, so exploration supplies practice in using the rules and
not only the rules themselves. \alienlogic{} inverts the pattern. O@$M_0$
alone reaches \result{96}--\result{100\%}, no Best@3 trajectory matches it, and
exploring first adds nothing (A4+O minus O@$M_0$ has a median of \result{$-0.2$}
points). \alienlogic{} is limited by discovery, and \aliencode{} by use.
Exploration can also interfere with use. Seed2.1 Pro scores lower under A4+O
than under O@$M_0$ (\result{56.7\%} against \result{62.9\%}).

\keepfinding
\findingbox{\findingcodeicon}{Finding 5. \aliencode{} hinges on two keystone
rules: systems that discover both correctly perform far better on subsequent
held-out tasks.}
Both keystone rules are positional, and together they enter \spec{51} of the
\spec{70} tasks. They are the index shift of \texttt{PLUCK} (R14) and the slice
shift of \texttt{CARVE} (R15). Of the \result{15} trajectories that end at or
above \result{50\%}, \result{13} state both correctly in their rule reports,
and of the \result{8} that end below \result{25\%}, \result{5} state neither.
Large accuracy jumps coincide with their discovery
(\cref{fig:keystone,case:keystone}). Of the \result{13} jumps of at least
\result{30} points between consecutive milestones, \result{12} occur in the
round in which the rule report newly states R15 correctly, and \result{11} in
the round in which it newly states R14. Across all \result{30} trajectories,
the number of rules stated correctly at $M_4$ correlates with $M_4$ at
$r=\result{0.85}$. This evidence is correlational, and the rule report never
enters the score.

\begin{diagnosticcase}[label=case:keystone, unbreakable, float=!tb]{One round finds the keystone rules}{GPT-5.6 Sol · \aliencode{}}{\findingcodeicon}
  \casemeta{GPT-5.6 Sol (\textsf{max})}{Trajectory 2, round 1, call 5}{Rule reports and held-out accuracy at $M_0$ and $M_1$}{12 tool calls in round 1}
  \begin{casepanel}{Probe as submitted, with the sandbox output (two further lines omitted)}
    \footnotesize
    \begin{tabular}{@{}l@{\qquad}l@{}}
      \texttt{SET xs AS STRAND(31,24,25,26)} & \\
      \texttt{EMIT(xs)} & \texttt{[1, 2, 3, 4]} \\
      \texttt{EMIT(STRAND(PLUCK(xs,27),PLUCK(xs,26),PLUCK(xs,25),PLUCK(xs,-28)))} & \texttt{[3, 2, 1, 4]} \\
      \texttt{EMIT(CARVE(xs,26,31))} & \texttt{[1, 2, 3]}
    \end{tabular}
  \end{casepanel}
  \begin{casepanel}{The same probe read with the two rules the system already reported at $M_0$}
    \footnotesize
    Integer literals are XOR-ed with 27 (R01) and \texttt{STRAND} reverses its
    arguments (R16), so \texttt{xs} is \texttt{[1, 2, 3, 4]} and the calls ask
    for \texttt{PLUCK} at indices 0, 1, 2, and $-1$ and for
    \texttt{CARVE(xs, 1, 4)}.\par\smallskip
    \begin{tabular}{@{}l@{\qquad}l@{\qquad}l@{}}
      \textbf{Call} & \textbf{Manual predicts} & \textbf{Sandbox returns} \\
      \texttt{PLUCK} at 0, 1, 2, $-1$ & \texttt{1, 2, 3, 4} & \texttt{4, 1, 2, 3} \\
      \texttt{CARVE(xs, 1, 4)} & \texttt{[2, 3, 4]} & \texttt{[1, 2, 3]}
    \end{tabular}
  \end{casepanel}
  \begin{casepanel}{Rule report}
    \footnotesize
    \begin{tabular}{@{}l@{\qquad}l@{\qquad}l@{}}
      & $M_0$ (8 of 31 rules correct) & $M_1$ (28 of 31 rules correct) \\
      R14 & \texttt{UNKNOWN} & \texttt{(get seq (- i 1))} \\
      R15 & \texttt{UNKNOWN} & \texttt{(slice seq (- lo 1) (- hi 1))}
    \end{tabular}
  \end{casepanel}
  \tcblower
  \small
  Every value the probe returns differs from the manual's, so one call
  exposes both keystone shifts. After this round the report states \result{28}
  of the \spec{31} rules, and held-out accuracy rises from \result{11.0\%} to
  \result{81.9\%}; on the \spec{51} tasks that use R14 or R15 it rises from
  \result{9.8\%} to \result{80.4\%}.
\end{diagnosticcase}

\begin{figure}[!tb]
  \centering
  \includegraphics[width=0.72\linewidth]{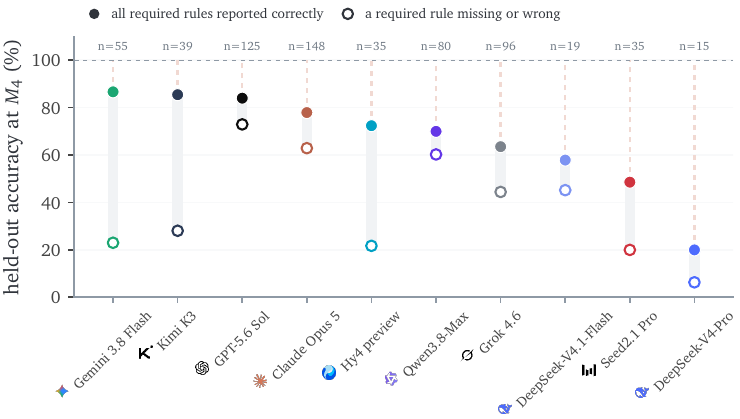}
  \caption{\textbf{Knowing a rule is not being able to use it.} For each
    system, pooled over its three \aliencode{} trajectories, filled dots give
  accuracy at $M_4$ on tasks whose required rules the $M_4$ rule report states
  correctly, and open dots on tasks with at least one required rule missing or
  stated incorrectly. The dashed segment up to 100\% is the remaining
  execution gap; $n$ counts task--trajectory pairs in the first group.}
  \label{fig:knowing-doing}
\end{figure}

\keepfinding
\findingbox{\findingcodeicon}{Finding 6. Knowing a rule does not guarantee
using it correctly.}
When a trajectory's $M_4$ rule report states every rule a task requires
correctly, the task is still solved only \result{73.4\%} of the time
(\result{647} task--trajectory pairs, \cref{fig:knowing-doing}). Two
trajectories state both keystone rules correctly yet end at \result{5.7\%} and
\result{12.9\%}. The dissociation also runs the other way. In \result{26} of
\result{30} trajectories, a rule stated correctly at one milestone later drops
out of the rule report, yet accuracy on the tasks that require it still rises,
from \result{11.4\%} to \result{16.9\%}. A task that requires a rule the
report states incorrectly is still solved \result{34.6\%} of the time. The rule
report is therefore a lossy readout of what a system can do and no substitute
for held-out accuracy. \Cref{case:scope} shows both
directions within one system.

\Needspace{0.16\textheight}
\subsubsection{RQ3. How reliable is exploration?}

\findingbox{\findingbothicons}{Finding 7. Same-system, same-budget trajectories
vary by tens of points, far beyond the noise from repeated answering.}
\begin{figure}[!tb]
  \centering
  \includegraphics[width=\linewidth]{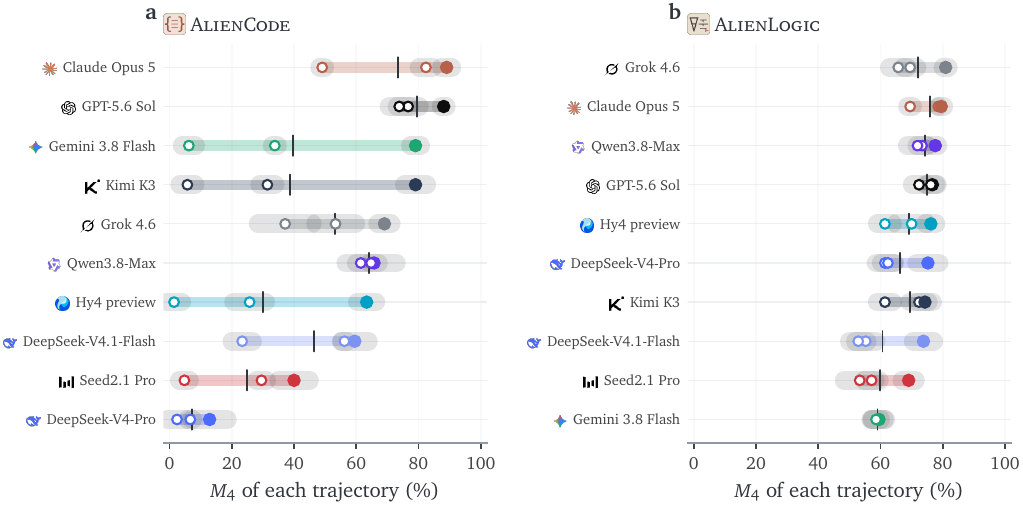}
  \caption{\textbf{Outcomes vary between trajectories, not between answers.} Each row
    is one system. Dots are its three trajectories' $M_4$ (filled for Best@3), the
  coloured band spans them, and the short vertical line marks Mean@3. The grey
  halo around each dot extends one standard deviation of that trajectory's
  three answers per question to either side.}
  \label{fig:run-spread}
\end{figure}
\begin{figure}[!tb]
  \centering
  \includegraphics[width=\linewidth]{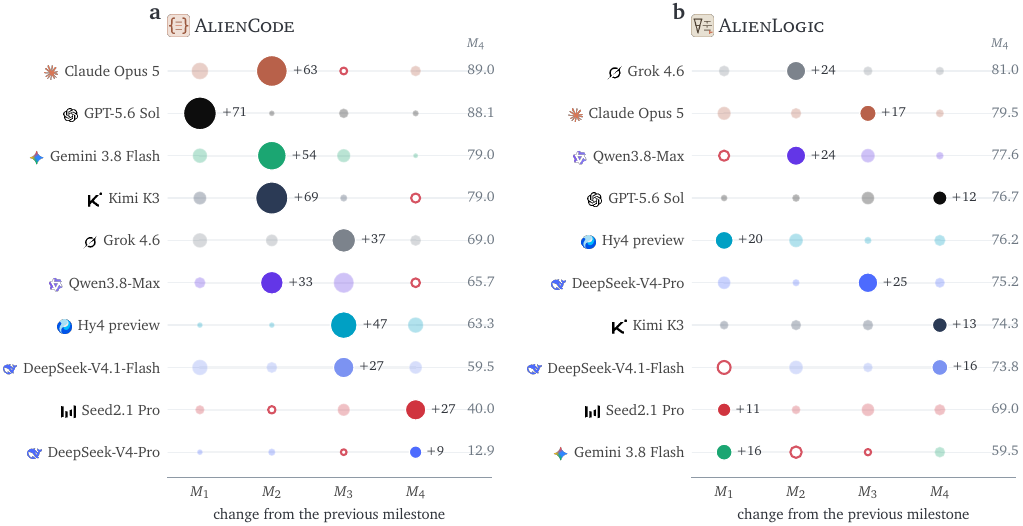}
  \caption{\textbf{When the gain arrives.} For each system's Best@3
  trajectory, bubble area is the change in held-out accuracy from the
  previous milestone ($M_t-M_{t-1}$). The solid bubble marks the largest step
  and is labelled with its size; a red ring marks a milestone at which
  accuracy fell.
  The right column gives $M_4$.}
  \label{fig:leap-timing}
\end{figure}
Answering noise is small (\cref{fig:run-spread}). Its standard deviation over
the three answering passes at the same milestone is at most \result{4.7} points
in \aliencode{} and \result{2.9} in \alienlogic{}. Trajectories of the same
system differ far more, by up to \result{73.3} points in \aliencode{} and
\result{20.9} in \alienlogic{}. In \aliencode{}, Kimi K3 ends between
\result{5.7\%} and \result{79.0\%} (\cref{case:scope}), and Gemini 3.8 Flash
between \result{6.2\%} and \result{79.0\%}. Every \aliencode{} trajectory
spends \result{45}--\spec{48} of its \spec{48} tool calls, so the spread does
not reflect a difference in effort. With outcomes this dispersed, the choice
of aggregate changes the ranking. Qwen3.8-Max ranks \result{sixth} in \aliencode{} by Best@3 and
\result{third} by Mean@3, while Kimi K3 falls from a tie for \result{third} to
\result{seventh}.

\keepfinding
\findingbox{\findingbothicons}{Finding 8. Exploration gains arrive in leaps, at
different times, and can reverse.}
In each Best@3 trajectory in \aliencode{}, the largest single step
contributes \result{47}--\result{92\%} of the retained gain
(\cref{fig:leap-timing}). The round in which that step arrives differs by
system. It is the first round for GPT-5.6 Sol, the second for Claude Opus 5,
Kimi K3, and Gemini 3.8 Flash, and the third for Grok 4.6 and Hy4 preview.
Seed2.1 Pro and DeepSeek-V4-Pro take it only in the last round and are still
rising when the budget ends. Continued exploration can also undo progress. Of
the \result{30} \aliencode{} trajectories, \result{6} end at least \result{3}
points below an earlier milestone, as does \result{1} of the \result{30}
\alienlogic{} trajectories. Gemini 3.8 Flash's Best@3 trajectory in
\alienlogic{} falls from \result{61.4\%} after the first round to
\result{54.3\%} after the second and ends at \result{59.5\%}
(\cref{case:refusal}).

\section{Discussion}
\label{sec:discussion}

\subsection{What Best@\texorpdfstring{$n$}{n} measures}
\label{sec:discussion-bestn}

Best@$n$ asks what endpoint a system can reach within $n$ independent
attempts. It answers whether a system can build an effective exploration
trajectory at all, but it is not a reliability measure: a system with one
strong trajectory and $n-1$ failures can outrank one whose every trajectory is
moderately strong. We therefore compare systems only at the same $n$ and report
Mean@$n$, every endpoint, and the best--worst range beside the ranking.

The best trajectory is chosen once, by its complete $M_4$ accuracy, and every
analysis of that system uses the same trajectory. Taking the best milestone
from one trajectory and the best answer to each task from another would
construct an outcome that no agent produced. For the same reason, Best@$n$ is
not pass@$n$, which takes the union of task-level successes and can exceed
every observed trajectory.

\subsection{Why the seed phase provides ten examples}
\label{sec:discussion-seed}

Before $M_0$, \aliencode{} gives every system ten fixed examples covering the
basic program forms used later: literals, operator calls, sequences, nesting,
and function bodies. This establishes a shared minimum fluency with the
language and prevents basic syntax learning from being conflated with
exploration, while leaving most hidden rules to be discovered in the four
scored rounds.

\section{Limitations}
\label{sec:limitations}

\bench{} deliberately reduces exploration to two deterministic, executable
worlds so that every experiment and held-out answer has an exact outcome. That
control is also its boundary: real scientific exploration involves noisy and
incomplete observations, costly or irreversible experiments, open-ended
hypothesis spaces, and horizons far longer than four rounds. Our results
therefore measure whether systems can acquire unfamiliar rules in verifiable
synthetic environments, not whether they can conduct real scientific
discovery. Three trajectories per system and three answers per question expose
trajectory and answering variability, but are insufficient to estimate either
distribution precisely.

\section{Conclusion}
\label{sec:conclusion}

We present \bench{}, a benchmark for how AI systems explore worlds whose rules
contradict familiar knowledge. Its two sandboxes contain \spec{55} discovery
targets and \spec{140} held-out tasks, and a program checks every answer.
Across ten frontier systems, four rounds of exploration lift the best
\aliencode{} trajectory to \result{89.0\%}, while the same turns without
environment feedback stay at or below \result{11.0\%}. Exploration works best
when the system designs its own experiments, but it is still far from
reliable. A system's exploration ability differs across tasks, stating a rule
correctly does not ensure using it, and trajectories of one system can end far
apart. \bench{} provides a testbed for studying exploration as a capability in
its own right and for developing systems that learn from their environments.
More broadly, the same design can evaluate exploration wherever unfamiliar
rules can be made executable and their consequences verified exactly, offering
insights for developing future AI systems for scientific discovery.

\fullauthorlist
\vspace{1.2\baselineskip}

\bibliographystyle{plainnat}
\bibliography{references}

\FloatBarrier
\appendix
\section*{Appendix}
\section{Experimental Details}
\label{app:protocol}

\subsection{Environments, tasks, and worked examples}

\bench{} contains two environments with the same exploration and evaluation
protocol but different executable objects. \aliencode{} exposes a small
programming language whose familiar-looking operators follow hidden semantics.
Its evaluator contains \spec{31} discovery targets and \spec{70} held-out tasks.
\alienlogic{} exposes a Fitch-style proof checker with \spec{24} discovery targets, each an active rule patch, and \spec{70} held-out tasks, including designated unprovable goals.
Programs must pass all private inputs, and proofs must verify. An unprovable
goal earns credit only for a correct refusal.

\aliencode{} includes deliberately counter-intuitive keywords and
\spec{8} red herrings. For example, \texttt{SHATTER} appears to mean destruction but
means multiply, while \texttt{EMIT(100)} prints \texttt{127}. Some rules are
scope-dependent, and no primitive performs addition, so the model must construct
addition from other operations. Engineering and algorithm tasks that require
code are evaluated on \spec{5} private legal inputs. Displayed examples specify
task interfaces rather than the private test set.

\alienlogic{} uses non-contiguous opaque rule IDs and side conditions, including
duplicate-premise and use-once guards that prevent a model from
discharging an assumption by simply restating a premise. Its designated
unprovable tasks are certified only within declared formula, line, indentation,
and node bounds.

Both environments use prescribed worked examples before $M_0$, but the
examples are sandbox-specific. \aliencode{} presents ten fixed worked
examples, balanced across five task bands (apply, interact, scope, engineer,
and algorithm). Each contains a task, a reference AlienCode program, and the
output the interpreter computes for it. The model submits no code in this
phase, so every trajectory reaches $M_0$ with identical evidence and no tool
calls charged. \alienlogic{} presents eight accepted worked proofs that show
valid proof syntax. $M_0$ is therefore comparable within a sandbox, but
comparisons of $M_4-M_0$ across sandboxes are only descriptive, because the
two sets of worked examples differ.

\subsection{Comparison with related benchmarks}

\Cref{tab:benchmark-comparison} places \bench{} beside the benchmarks of
\cref{sec:related}: where each one's evidence comes from, how it controls
novelty, and where it finally tests competence.

\begin{table}[!tbp]
  \centering
  \caption{\textbf{Where the evidence comes from, and where competence is
  tested.} Each row states a benchmark's primary protocol rather than ranking
  design choices. \bench{} measures exploration by combining autonomous
  probe selection, milestone
  measurement, and unseen-task evaluation after interaction ends; no single
  column alone defines that distinction.}
  \label{tab:benchmark-comparison}
  \scriptsize
  \setlength{\tabcolsep}{3pt}
  \renewcommand{\arraystretch}{1.15}
  \begin{tabularx}{\linewidth}{@{}
    >{\raggedright\arraybackslash}p{2.35cm}
    Y{1.14} Y{1.00} Y{1.10} Y{0.92} Y{0.98} Y{0.86}@{}}
    \toprule
    & & \multicolumn{2}{c}{\textit{Learning setup}}
      & \multicolumn{3}{c@{}}{\textit{What is measured}} \\
    \cmidrule(lr){3-4}\cmidrule(l){5-7}
    \textbf{Benchmark} & \textbf{Target} & \textbf{Evidence}
      & \textbf{Novelty control} & \textbf{Progress} & \textbf{Transfer}
      & \textbf{Ground truth} \\
    \midrule
    CL-bench \citep{dou2026clbench}
                   & Context learning         & Provided context      & Expert-authored content & Final score          & Same context         & Expert rubrics, LLM verifier \\
    EvaLearn \citep{dou2025evalearn}
                   & Sequential learning      & Prior solved tasks    & Authored task sequences & Learning curve       & Later related tasks  & Rubrics, LLM verifier \\
    SE-Bench \citep{yuan2026sebench}
                   & Weight internalization   & Docs, training tasks  & Obfuscated APIs         & Pre/post score       & Closed-book held-out & Tests, AST checks \\
    SWE-bench \citep{jimenez2024swebench}
                   & Software repair          & Issue, codebase       & Real GitHub issues      & Final patch          & Same repository      & Test suites \\
    DiscoveryWorld \citep{jansen2024discoveryworld}
                   & Scientific investigation & Agent actions         & Fictional worlds        & Final score          & Same world           & World state \\
    NewtonBench \citep{zheng2026newtonbench}
                   & Physical-law discovery   & Chosen experiments    & Counterfactual laws     & Final equation       & Inferred law         & Symbolic equivalence \\
    EdgeBench \citep{zhu2026edgebench}
                   & Long-horizon learning    & Environment feedback  & New real-world tasks    & Learning curve       & Same task            & Task-specific evaluator \\
    \midrule
    \rowcolor{black!7}
    \shortstack[l]{\textbf{\textsc{Exploration}}\\
                   \textbf{\textsc{Bench} (ours)}}
                   & \textbf{Exploration}
                                            & \textbf{Chosen probes and feedback}
                                                                     & \textbf{Executable rules that conflict with priors}
                                                                                               & $\bm{M_0\!\to\!M_4}$ \textbf{milestones}
                                                                                                                      & \textbf{Unseen tasks after interaction}
                                                                                                                                             & \textbf{Interpreter; proof checker, bounded certifier} \\
    \bottomrule
  \end{tabularx}
\end{table}

\subsection{Models and reasoning controls}
\label{app:provenance}

We evaluate GPT-5.6 Sol, Claude Opus 5, Qwen3.8-Max (0902),
Gemini 3.8 Flash, DeepSeek-V4.1-Flash, DeepSeek-V4-Pro, Grok 4.6, Kimi K3,
Seed2.1 Pro (0915), and Hy4 preview. Each system uses the highest reasoning
setting its API offers, fixed across trajectories and sandboxes; the
\textsf{high}/\textsf{xhigh}/\textsf{max} labels name request settings, not
comparable compute levels. Dated suffixes identify the exact model versions
evaluated.

Each system runs $n{=}3$ independent trajectories per sandbox. Interactions use
structured tool calls: the environment executes each call and returns a tool
result (\cref{app:formal-protocol}).

\subsection{Interaction budget and control conditions}
\label{app:controls}

Exploration keeps one continuous history. After each round, the system is
tested in a discarded, tool-disabled copy of the conversation, so testing never
becomes evidence for later rounds.

\Needspace{5\baselineskip}
Milestone number is a protocol stage, not a resource unit. At each milestone we
record cumulative tool calls $C$, probe units $P$, and exploration
tokens $T$. Each of the four rounds allows at most \spec{12} tool calls in
\aliencode{} and \spec{12} proofs in \alienlogic{}, the latter capped at
\spec{48} across the run. \aliencode{} packs up to
five probe units into one tool call, charging one for each top-level
\texttt{EMIT} argument and three for each such observation inside a loop body;
\alienlogic{}
submits one verifier-evaluated proof per call, so $C=P$. We use actual
consumption, not the caps, in every budget we report.
The loop charge is fixed rather than multiplied by the realized iteration
count, and a call over the five-unit cap is rejected in full and executes no
code. Feedback is capped separately at 15 lines and 600 characters per call,
and returned output never adds probe units. The worked examples are context
rather than model calls and are not charged. The budget is recorded and
reported but never scored.

\newcommand{\BudgetRows}{%
\modelname{claude-ai}{Claude Opus 5} & 48 & 168 & 11.59 & 48 & 0.67 \\
\modelname{openai}{GPT-5.6 Sol} & 48 & 131 & 3.62 & 48 & 0.91 \\
\modelname{gemini-color}{Gemini 3.8 Flash} & 45 & 124 & 2.41 & 48 & 1.08 \\
\modelname{kimi}{Kimi K3} & 45 & 216 & 2.01 & 48 & 0.38 \\
\modelname{grok}{Grok 4.6} & 48 & 197 & 0.50 & 48 & 0.15 \\
\modelname{qwen-color}{Qwen3.8-Max} & 48 & 162 & 0.99 & 48 & 0.68 \\
\modelname{hunyuan-color}{Hy4 preview} & 48 & 210 & 1.41 & 48 & 0.33 \\
\modelname{deepseek-color}{DeepSeek-V4.1-Flash} & 48 & 143 & 10.85 & 48 & 1.63 \\
\modelname{bytedance-light}{Seed2.1 Pro} & 48 & 219 & 0.34 & 48 & 0.18 \\
\modelname{deepseek-color}{DeepSeek-V4-Pro} & 48 & 210 & 4.55 & 48 & 0.37 \\
}

\Cref{tab:budget} reports the budget of each system's Best@3 trajectory. Every
trajectory spends nearly all of its tool calls, while exploration tokens differ
by more than an order of magnitude between systems.

\begin{table}[!ht]\centering
  \caption{\textbf{Exploration budget of each system's Best@3 trajectory.}
  $C$ is tool calls, $P$ probe units, and $T$ exploration tokens in millions,
  summed over the four rounds. In \alienlogic{}, $C=P$. Rows follow the
  \aliencode{} order of \cref{tab:main}.}
  \label{tab:budget}
  \small
  \begin{tabular}{@{}l rrr rr@{}}
    \toprule
    & \multicolumn{3}{c}{\aliencode{}} & \multicolumn{2}{c@{}}{\alienlogic{}} \\
    \cmidrule(lr){2-4}\cmidrule(l){5-6}
    System & {$C$} & {$P$} & {$T$ (M)} & {$C=P$} & {$T$ (M)} \\
    \midrule
    \BudgetRows
    \bottomrule
  \end{tabular}
\end{table}

\subsection{Formal tool-call protocol}
\label{app:formal-protocol}

\begin{algorithm}[!tb]
\small
\caption{The \bench{} protocol for structured tool calls. Fixed worked examples
  establish $M_0$. After each round the benchmark records $C$, $P$, and $T$, and
  the system is tested in discarded, tool-disabled copies of the conversation,
  so only probe feedback updates $H$.}
\label{alg:protocol}
\begin{algorithmic}[1]
\Require environment tool $\mathcal E$ (hidden rules $\mathcal R$); manual $\mathcal M$; fixed worked examples $\mathcal D$; held-out set $\mathcal T$; rounds $N$; sandbox caps $\bar{\mathbf B}$
\Ensure milestone scores $(M_0,\dots,M_N)$, rule reports
  $(S_0,\dots,S_N)$, and actual budgets
  $(\mathbf B_0,\dots,\mathbf B_N)$
\State $H \gets \textsc{InitHypothesis}(\mathcal M,\mathcal D)$
  \Comment{identical evidence; no model tool calls}
\For{$i \gets 0$ \textbf{to} $N$}
    \State $\mathbf B_i \gets \textsc{CumulativeUsage}(H)$
    \State $S_i \gets
      \textsc{ReportRules}(\textsc{ToolDisabledCopy}(H))$
    \State $M_i \gets \textsc{ScoreToolDisabledCopies}(H,\mathcal{T})$
    \If{$i < N$} \Comment{\textbf{Evidence acquisition}. Run probes and update}
        \While{the model requests a call to $\mathcal E$ and budget remains}
            \State $x \gets \textsc{ToolArguments}(H)$;\;
              $f \gets \textsc{Execute}(\mathcal E,x)$
            \State $H \gets \textsc{ReturnToolResult}(H,x,f)$
        \EndWhile
    \EndIf
\EndFor
\State \Return $(M_0,\dots,M_N)$, $(S_0,\dots,S_N)$, $(\mathbf B_0,\dots,\mathbf B_N)$
\end{algorithmic}
\end{algorithm}

Each request continues the API's structured conversation history, keeping
tool-call IDs and reasoning-state fields instead of flattening them into text,
so later probes condition on the full interaction history.

For trajectory $r$, the cumulative exploration budget at milestone $t$ is
\begin{equation}
  \mathbf B_{r,t}
  =\bigl(C_{r,t},P_{r,t},T_{r,t}\bigr),
  \label{eq:actual-budget}
\end{equation}
where $C$ is executed tool calls and therefore submitted probes, $P$ is charged
probe units, and $T$ is API-reported input plus output tokens for model
requests made during exploration.

\subsection{Executable outcomes and metric definitions}
\label{app:formal-metrics}

Let $a_{\tau,t}=A(\tau;\Hstate)$ be the response produced for unseen task
$\tau$ in an independent tool-disabled copy of the conversation at milestone $t$. The benchmark first
reduces every response to an executable binary outcome
\begin{equation}
  y_{\tau,t}=V_{\mathcal E}(\tau,a_{\tau,t})\in\{0,1\},
  \label{eq:task-outcome}
\end{equation}
where $V_{\mathcal E}$ is the sandbox-specific verifier. \aliencode{} holds two
kinds of task. For the \spec{69} synthesis tasks the model submits a program,
which passes only if it is correct on every evaluator input; for the
\spec{1} prediction task the task supplies the program $z_\tau$ and the
model submits the output it expects, which passes only on an exact match with
what the interpreter actually prints:
\begin{equation}
  V_{\mathcal E}^{\textsc{Code}}(\tau,a)=
  \begin{cases}
    \displaystyle\prod_{z\in\mathcal{Z}_\tau}
      \mathbb{1}\!\left[
        \operatorname{Exec}_{\mathcal E}(a,z)=g_\tau^{\mathcal E}(z)
      \right], & \text{synthesis},\\[6pt]
    \mathbb{1}\!\left[
      a=\operatorname{Exec}_{\mathcal E}(z_\tau)
    \right], & \text{prediction}.
  \end{cases}
  \label{eq:code-verifier}
\end{equation}
Both branches compare normalized text, so trailing whitespace and an
equal-valued numeric literal do not decide a verdict.
For \alienlogic{}, let $u_\tau=1$ denote a theorem designated unprovable within
the benchmark's declared certifier bounds. Then
\begin{equation}
  V_{\mathcal R}^{\textsc{Logic}}(\tau,a)=
  \begin{cases}
    \mathbb{1}\!\left[\operatorname{ProofCheck}_{\mathcal R}(\tau,a)
      =\textsc{Accept}\right], & u_\tau=0,\\
    \mathbb{1}\!\left[\operatorname{Decline}(a)\right], & u_\tau=1.
  \end{cases}
  \label{eq:logic-verifier}
\end{equation}

For trajectory $r$ at milestone $t$, held-out accuracy $M_{r,t}$ is the
three-answer mean of \cref{eq:milestone-accuracy} over the full scored set
$\mathcal{T}$. The per-pass scores $M^{(k)}_{r,t}$ give the answering noise
$\sigma_{r,t}$. The \emph{step} $s_{r,t}=M_{r,t}-M_{r,t-1}$ is the change
during round $t$, and the \emph{retained gain} is their sum,
\begin{equation}
  G_r = M_{r,4}-M_{r,0}=\sum_{t=1}^{4} s_{r,t}.
  \label{eq:retained-gain}
\end{equation}
We describe how a trajectory reaches its endpoint by the share of $G_r$ carried
by its largest step and by whether it ends below an earlier milestone. These
quantities describe the selected trajectory and are never used to select it.

The rule report score of \cref{sec:metrics} scores the reported rule set $S_t$
against the evaluator-side \aliencode{} rule inventory
$\mathcal{R}_{\mathrm{eval}}$ as the number of rules it states correctly:
\begin{equation}
  \mathrm{RuleReport}_t
  =\sum_{\rho\in\mathcal{R}_{\mathrm{eval}}}J(\rho,S_t),
  \label{eq:rule-report}
\end{equation}
It is elicited at every milestone; we report its value at $M_4$.

Over $n$ trajectories, the model score and its selected trajectory are
\begin{equation}
  \operatorname{Best@}n=\max_{r}M_{r,4},
  \qquad
  r^\star=\min\arg\max_r M_{r,4}.
  \label{eq:run-aggregation}
\end{equation}
All trajectory-level quantities paired with the headline score---$M_0$,
$G$, intermediate milestones, rule reports, budgets, and errors---are read
from $r^\star$. We separately report
$\operatorname{Mean@}n=n^{-1}\sum_r M_{r,4}$, every endpoint, and the
best--worst range, and rank systems only at the same $n$. We attach no
confidence interval to Best@$n$: a maximum depends explicitly on $n$, which is
why $n$ and every trajectory are reported beside it.

The five exploration conditions differ only in how evidence is obtained; the
sixth, open-book answering, supplies the complete rule set as the reference used
in RQ2.
\begin{itemize}[leftmargin=1.4em,itemsep=2pt]
  \item Direct answering ($d$). No model turns occur after $M_0$.
  \item Without-tool answering ($w$). Model turns for
        hypothesis revision occur without environment feedback.
  \item Fixed-probe exploration ($f$). Probes are
        sampled deterministically from a fixed pool of grammar-valid templates.
        One sequence serves all ten systems: its seed is fixed, and its
        per-call volume follows one mid-ranked trajectory, the Best@3 trajectory
        of Grok 4.6, so every system meets the same probes, independent of its
        own hypotheses.
  \item Hindsight exploration ($h$). The probes of the same system's Best@3
        trajectory in that sandbox, chosen after the fact by its $M_4$, are
        replayed through the same tool-calling interface.
  \item Autonomous exploration ($a$). Probes are selected from the
        trajectory's own history.
  \item Open-book answering ($o$). The complete rule set is supplied before
        evaluation, either before exploration (O@$M_0$) or after the four
        autonomous rounds (A4+O).
\end{itemize}
Hindsight replay keeps the structured exchange of tool calls and tool results
and changes only who chooses the probe. The replayed probes are submitted
verbatim but executed live, so their feedback is recomputed rather than copied,
and the system still records its own hypotheses between probes. The difference
between autonomous and hindsight exploration therefore reflects choosing the
probes, not a change of interaction protocol. Because the replayed sequence
comes from the system's best trajectory rather than an average over
trajectories, the comparison is conditioned on that particular sequence.

\section{Per-system results}
\label{app:per-system}

\begin{figure}[p]
  \centering
  \includegraphics[width=\linewidth,height=0.90\textheight,keepaspectratio]{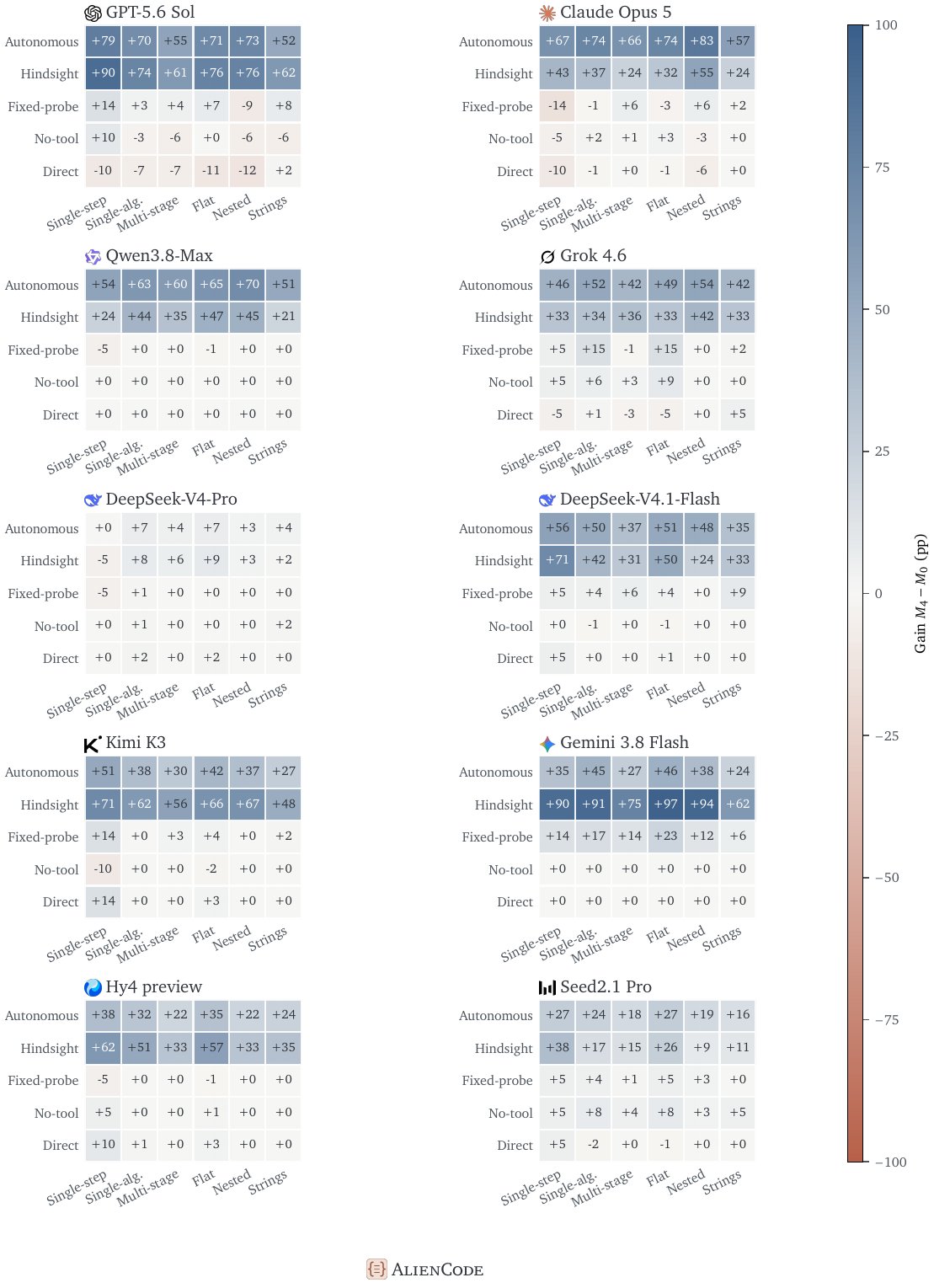}
  \caption{\textbf{Task-family gains for every \aliencode{} system.} Cells
  show $M_4-M_0$ within each task family, in percentage points. Tasks are
  split by compositional depth and by representation, so each task enters one
  column of each. Autonomous rows average three trajectories; control rows
  are one trajectory each.}
  \label{fig:profiles-code-all}
\end{figure}

\begin{figure}[p]
  \centering
  \includegraphics[width=\linewidth,height=0.90\textheight,keepaspectratio]{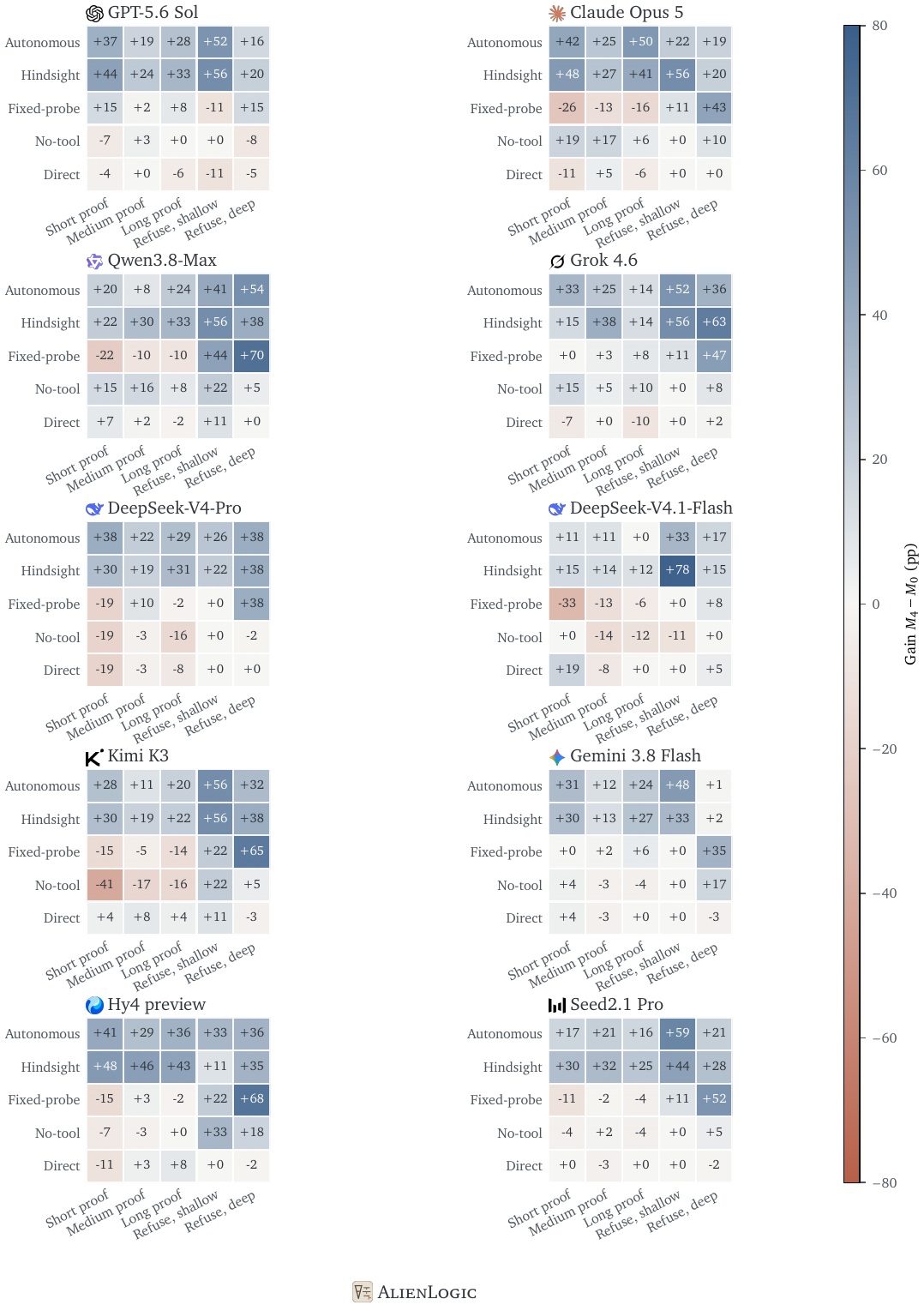}
  \caption{\textbf{Task-family gains for every \alienlogic{} system.} Cells
  show $M_4-M_0$ within each task family, in percentage points. Columns are
  the five families of \cref{fig:task-composition}: provable theorems by proof
  length, unprovable ones by how deeply the goal nests. Autonomous rows
  average three trajectories; control rows are one trajectory each.}
  \label{fig:profiles-logic-all}
\end{figure}

\begin{figure}[p]
  \centering
  \includegraphics[width=\linewidth,height=0.92\textheight,keepaspectratio]{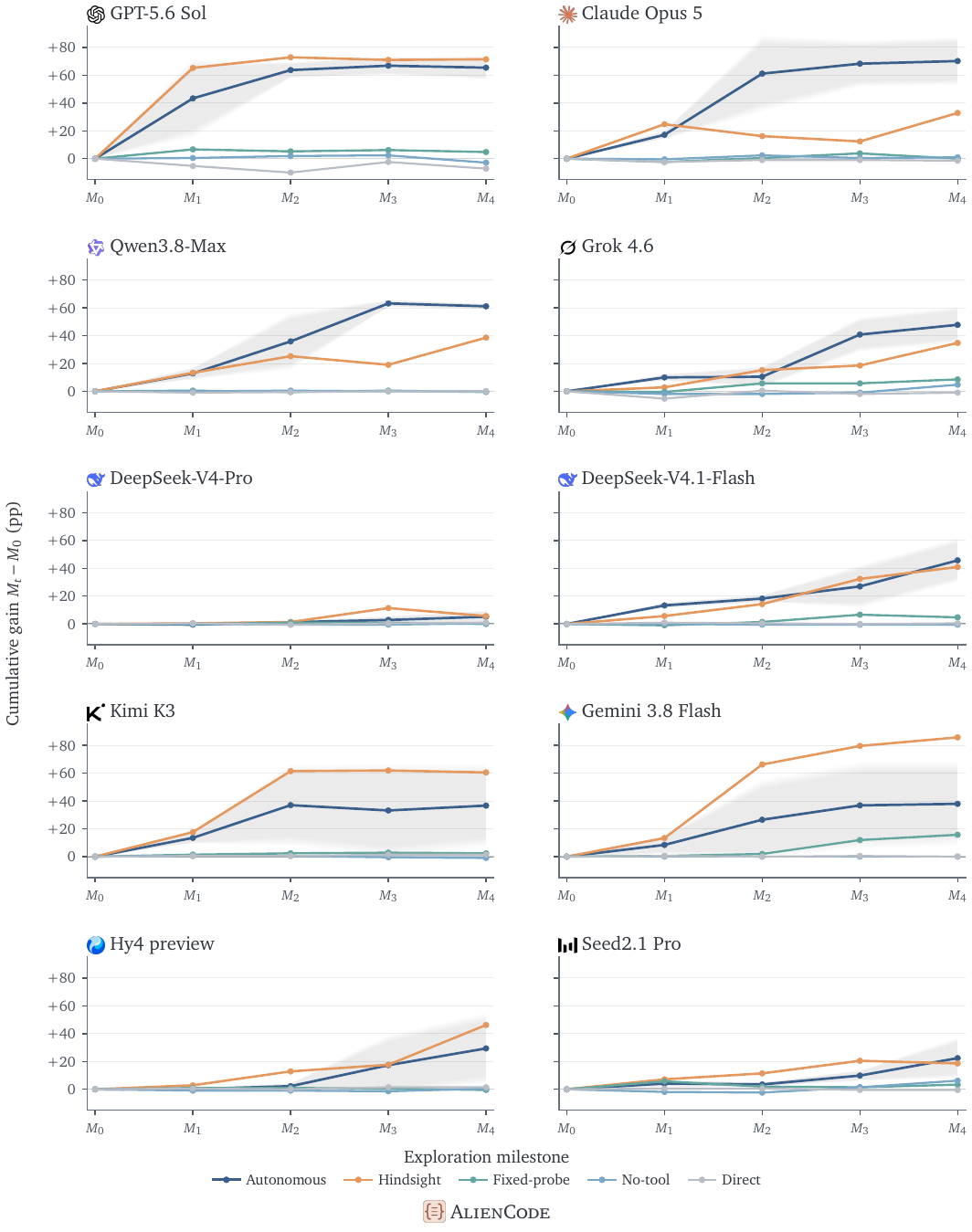}
  \caption{\textbf{Exploration trajectories for every \aliencode{} system.}
  Curves show cumulative gain $M_t-M_0$. The autonomous curve is the mean of
  three trajectories, and its shading is $\pm$ population SD across them;
  each control curve is a single trajectory.}
  \label{fig:trajectories-code-all}
\end{figure}

\begin{figure}[p]
  \centering
  \includegraphics[width=\linewidth,height=0.92\textheight,keepaspectratio]{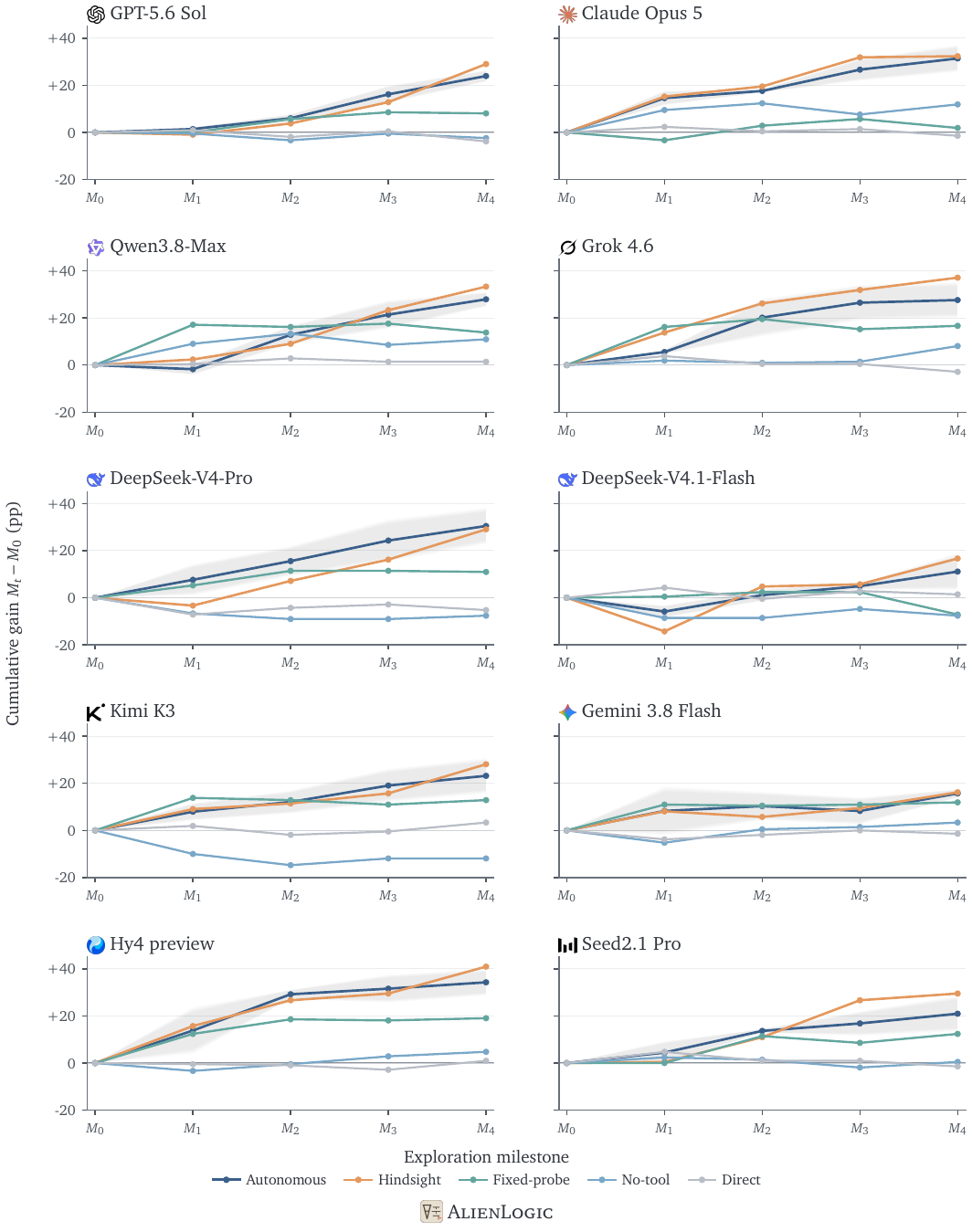}
  \caption{\textbf{Exploration trajectories for every \alienlogic{} system.}
  Curves show cumulative gain $M_t-M_0$. The autonomous curve is the mean of
  three trajectories, and its shading is $\pm$ population SD across them;
  each control curve is a single trajectory.}
  \label{fig:trajectories-logic-all}
\end{figure}

\Cref{fig:profiles-code-all,fig:profiles-logic-all,fig:trajectories-code-all,fig:trajectories-logic-all}
break each system's gain down by task family and by milestone under the five
exploration conditions. Autonomous exploration averages three trajectories;
each control condition contributes the one trajectory scored for the control
study. Scores are three-answer means (\cref{eq:milestone-accuracy}), and gains
are measured from each condition's own $M_0$.

\clearpage
\section{Case studies}
\label{app:cases}

Each case follows one trajectory to show a mechanism behind a finding; the
cases are illustrations, not further evidence. Programs, proofs, and rule
reports are shown as logged. Explanations the systems wrote in Chinese, the
language of the prompts, are translated.

\begin{diagnosticcase}[label=case:scope]{Same system, same budget, one scope rule apart}{Kimi K3 · \aliencode{}}{\findingcodeicon}
  \casemeta{Kimi K3 (\textsf{max})}{Trajectories 2 and 3; task A46 at $M_4$}{Rule reports, held-out accuracy, $M_4$ programs}{48 and 45 tool calls}
  \begin{casepanel}{Rules stated correctly (of 31) and held-out accuracy}
    \footnotesize
    \begin{tabular}{@{}l@{\qquad}rr@{\qquad}rr@{}}
      & \multicolumn{2}{l}{Trajectory 2} & \multicolumn{2}{l}{Trajectory 3} \\
      $M_0$ & 7 & \result{1.4\%} & 3 & \result{2.4\%} \\
      $M_1$ & 10 & \result{12.9\%} & 12 & \result{12.4\%} \\
      $M_2$ & 14 & \result{15.2\%} & 27 & \result{81.0\%} \\
      $M_3$ & 19 & \result{7.6\%} & 28 & \result{82.9\%} \\
      $M_4$ & 22 & \result{5.7\%} & 28 & \result{79.0\%}
    \end{tabular}\par\smallskip
    R14 (\texttt{PLUCK} index shift) is stated from $M_4$ in trajectory 2 and
    from $M_2$ in trajectory 3. R01+ (integer literals inside a function body
    are XOR-ed with 53; \spec{69} of the \spec{70} tasks need it) is stated
    only at $M_0$ in trajectory 2 and from $M_3$ in trajectory 3.
  \end{casepanel}
  \begin{casepanel}{Task A46 at $M_4$: write \texttt{CRAFT last(lst)} that returns the last element}
    \footnotesize
    \begin{tabular}{@{}l@{\qquad}l@{\qquad}l@{}}
      Trajectory 2 & \texttt{DELIVER PLUCK(lst, 27)} & \failstep{} \texttt{IndexError} on every test \\
      Trajectory 3 & \texttt{DELIVER PLUCK(lst, 53)} & \passstep{}
    \end{tabular}\par\smallskip
    Trajectory 2 explains its answer: ``\texttt{PLUCK(seq, i)} is
    \texttt{seq[i-1]}, so $i=0$ returns the last element; integer literals are
    XOR-ed with 27, so 0 is written 27.''
  \end{casepanel}
  \tcblower
  \small
  Both trajectories state the index shift at $M_4$, and trajectory 2
  states 22 rules. It encodes the literal with the top-level rule inside a
  function body, where 27 becomes 46, so the index runs out of range. On the
  \spec{51} tasks that use R14 or R15, trajectory 2 solves \result{2.6\%} at
  $M_4$ and trajectory 3 \result{75.8\%}. The report also lags behind use:
  trajectory 3 reaches \result{81.0\%} at $M_2$ before it states R01+.
\end{diagnosticcase}

\begin{diagnosticcase}[label=case:refusal]{A correct refusal overwritten by a later belief}{Gemini 3.8 Flash · \alienlogic{}}{\findinglogicicon}
  \casemeta{Gemini 3.8 Flash (\textsf{high})}{Trajectory 2 (Best@3); theorem T50}{Answers at $M_1$ and $M_2$, rule reports, held-out accuracy}{Four exploration rounds}
  \begin{casepanel}{Theorem T50, designated unprovable: a patched rule lets each premise be cited only once}
    \footnotesize
    Premise \texttt{AND(p, NOT(p))}; goal \texttt{NOT(r)}.
  \end{casepanel}
  \begin{casepanel}{Answers}
    \footnotesize
    $M_1$, correct in all three answers \passstep{}: ``\texttt{ANSWER:
    UNPROVABLE}. Under K28 and K24, the single premise
    \texttt{AND(p, NOT(p))} cannot serve both \texttt{AND\_E1} and
    \texttt{AND\_E2}, so the contradiction between \texttt{p} and
    \texttt{NOT(p)} cannot be derived.''
    \par\smallskip
    $M_2$, wrong in all three answers, as at $M_3$ and $M_4$ \failstep{}:\par
    \begin{tabular}{@{}lll@{\qquad\qquad}lll@{}}
      \texttt{1.} & \texttt{AND(p, NOT(p))} & \texttt{[premise]} & \texttt{5.} & \texttt{| NOT(p)} & \texttt{[AND\_E2 3]} \\
      \texttt{2.} & \texttt{| r} & \texttt{[assume]} & \texttt{6.} & \texttt{| BOT} & \texttt{[NOT\_E 4, 5]} \\
      \texttt{3.} & \texttt{| AND(p, NOT(p))} & \texttt{[reit 1]} & \texttt{7.} & \texttt{NOT(r)} & \texttt{[NOT\_I 2-6]} \\
      \texttt{4.} & \texttt{| p} & \texttt{[AND\_E1 3]} & & &
    \end{tabular}
  \end{casepanel}
  \begin{casepanel}{The rule report's entry on premise use}
    \footnotesize
    \begin{tabular}{@{}l@{\quad}p{0.84\linewidth}@{}}
      $M_0$ & ``A premise line cannot be directly cited more than once without
              explicitly reiterating it via reit.'' \\
      $M_2$ & ``Each declared top-level premise can be directly referenced at most
              once across the entire proof script.'' \\
      $M_4$ & ``A declared premise can only be directly referenced at most once (can
              be bypassed by copying via a single reit into a reusable derived line).''
    \end{tabular}
  \end{casepanel}
  \tcblower
  \small
  At $M_1$ the system refuses correctly. From $M_2$ it routes around the
  restriction by reiterating the premise, which the checker rejects because a
  conjunction cannot be reiterated (K06), and by $M_4$ its report states the
  workaround as a rule. The workaround holds for an atomic premise, and on
  theorem H13, which shares one atomic premise between two implications, the
  checker accepts it. Accuracy falls from \result{61.4\%} at $M_1$ to
  \result{54.3\%} at $M_2$: \result{five} theorems answered correctly in all
  three answers at $M_1$ fail in all three at $M_2$, \result{four} of them
  designated unprovable, and \result{three} move the other way.
\end{diagnosticcase}

\begin{diagnosticcase}[label=case:nofeedback]{Thinking without evidence withdraws what the examples showed}{Kimi K3 · \aliencode{}}{\findingcodeicon}
  \casemeta{Kimi K3 (\textsf{max})}{Without-tool answering}{Rule reports at $M_0$ to $M_4$; round-4 turn}{Four rounds of model turns, no tool calls}
  \begin{casepanel}{Rule report}
    \footnotesize
    Rules stated correctly: 16, 3, 7, 3, and 1 at $M_0$ through $M_4$; entries
    marked \texttt{UNKNOWN} rise from 9 to 27. Held-out accuracy moves from
    \result{1.4\%} to \result{0.5\%}.\par\smallskip
    \begin{tabular}{@{}l@{\qquad}l@{\qquad}l@{}}
      & $M_0$ & $M_4$ \\
      R01 & \texttt{(\^{} n 27)} & \texttt{UNKNOWN} \\
      R01+ & \texttt{(\^{} n 53)} & \texttt{UNKNOWN} \\
      R22 & \texttt{(+ (len seq) 1)} & \texttt{UNKNOWN}
    \end{tabular}
  \end{casepanel}
  \begin{casepanel}{Round-4 turn (excerpt, translated)}
    \footnotesize
    ``No new evidence; the posterior is not updated relative to the last round.
    \ldots{} Integer values are unstable: 32 to 59, 9 to 60, list elements 4/5
    to 48/49, length 3 to 4; no reliable uniform offset.''
  \end{casepanel}
  \tcblower
  \small
  The observations it cites are the ones its $M_0$ report explained:
  32 to 59 is 32 XOR 27, 9 to 60 is 9 XOR 53, and a length of 3 read as 4 is
  R22. With no new evidence, the extra turns make the report more cautious
  rather than more accurate, and the worked examples still in context stop
  being trusted.
\end{diagnosticcase}

\end{document}